\documentclass[10pt,journal,compsoc]{IEEEtran}
\ifCLASSOPTIONcompsoc
  \usepackage[nocompress]{cite}
\else
  \usepackage{cite}
\fi
\ifCLASSINFOpdf
   \usepackage[pdftex]{graphicx}
\else
\fi
\usepackage{amsmath}
\usepackage{algorithmic}

\ifCLASSOPTIONcaptionsoff
 \usepackage[nomarkers]{endfloat}
\let\MYoriglatexcaption\caption
\renewcommand{\caption}[2][\relax]{\MYoriglatexcaption[#2]{#2}}
\fi
\usepackage{url}

\usepackage{helvet}  
\usepackage{courier}  
\usepackage{url}  
\usepackage{epsfig}
\usepackage{graphicx}
\usepackage{amsmath}
\usepackage{amssymb}

\usepackage{color}
\usepackage{multirow}

\newcommand{\edit}[1]{#1}

\definecolor{light-gray}{gray}{0.95}    
\definecolor{orange}{rgb}{1,0.5,0}      
\definecolor{pinkr}{RGB}{255,140,197}
\definecolor{bluer}{RGB}{86,193,255}

\begin{document}
%
\title{Progressive and Aligned Pose Attention Transfer for Person Image Generation}
%
%
%
%

\author{Zhen~Zhu*\thanks{*Authors contribute equally.},
        Tengteng~Huang*,
        Mengde~Xu,
        Baoguang~Shi, 
        Wenqing~Cheng, \\
        Xiang~Bai$^\dagger$\thanks{$^\dagger$Corresponding author.}~\IEEEmembership{Senior Member,~IEEE}
\IEEEcompsocitemizethanks{
\IEEEcompsocthanksitem Z. Zhu is with the Department of Computer Science, University of Illinois, Urbana Champaign, IL, 61801. \protect \\
E-mail: zhenzhu4@illinois.edu
\IEEEcompsocthanksitem T. Huang, M. Xu, W. Cheng are with the School of Electronic Information and Communications, Huazhong University of Science and Technology, Wuhan, 430074, China. \protect \\
E-mail: \{huangtengtng, mdxu, chengwq\}@hust.edu.cn.
\IEEEcompsocthanksitem B. Shi is now a Senior Researcher at Microsoft, Redmond. \protect \\
E-mail: shibaoguang@gmail.com.
\IEEEcompsocthanksitem X. Bai is with the School of Artificial Intelligence and Automation, Huazhong University of Science and Technology, Wuhan, 430074, China. \protect \\
E-mail: xbai@hust.edu.cn.
}}
\IEEEtitleabstractindextext{%

\begin{abstract}
This paper proposes a new generative adversarial network for pose transfer, \emph{i.e.}, transferring the pose of a given person to a target pose. We design a progressive generator which comprises a sequence of transfer blocks.
Each block performs an intermediate transfer step by modeling the relationship between the condition and the target poses with attention mechanism.
Two types of blocks are introduced,
namely Pose-Attentional Transfer Block~(PATB) and Aligned Pose-Attentional Transfer Block~(APATB).
Compared with previous works, our model generates more photorealistic person images that retain better appearance consistency and shape consistency compared with input images.
We verify the efficacy of the model on the Market-1501 and DeepFashion datasets, using quantitative and qualitative measures.
Furthermore, we show that our method can be used for data augmentation for the person re-identification task, alleviating the issue of data insufficiency.
Code and pretrained models are available at: \url{https://github.com/tengteng95/Pose-Transfer.git}.
\end{abstract}


}

\maketitle

\IEEEdisplaynontitleabstractindextext

%
\IEEEpeerreviewmaketitle


\IEEEraisesectionheading{\section{Introduction}\label{sec:introduction}}

    \IEEEPARstart{P}{erson} image generation is of great value in many real-world applications, such as virtual try-on \cite{viton}, video generation with a sequence of poses \cite{videogenerating} and data augmentation for person re-identification \cite{unlabeledReID}. Specifically in this paper, we focus on the \emph{pose transfer} task, aiming at transferring a person from one pose to another, as first introduced in \cite{poseguided}.
	
\begin{figure}
		\centering
		\includegraphics[width=0.48\textwidth]{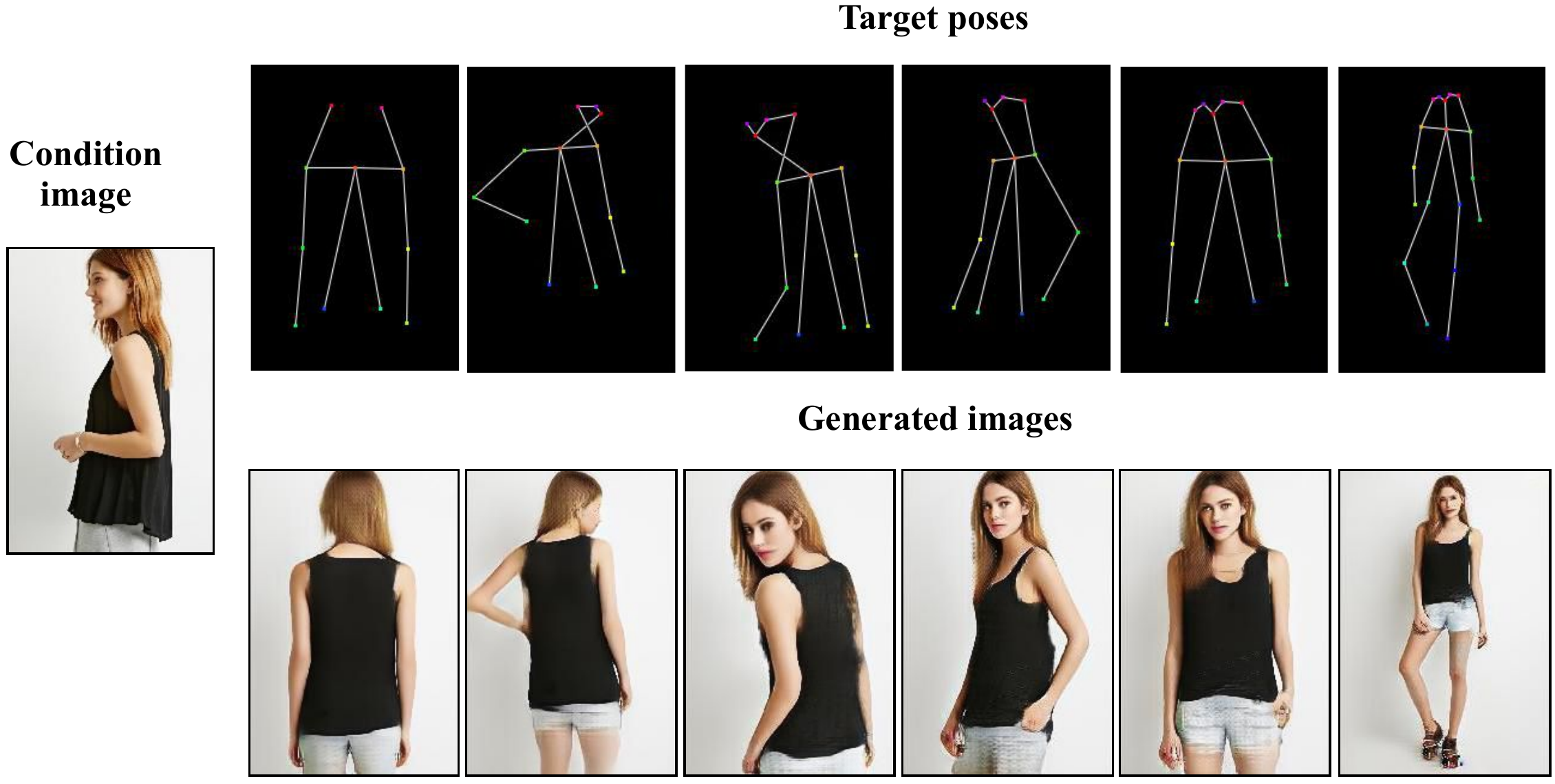}
		\caption{Generated examples by our method based on different target poses. Zoom in for better details.} \label{fig:introduction}
	\end{figure}

	Pose transfer is exceptionally challenging, particularly when given only the partial observation of the person. 
	As exemplified in Fig.~\ref{fig:introduction}, the generator needs to infer the unobserved body parts in order to generate plausible images. 
	Many previous work~\cite{poseguided,DeformableGAN,Disentangled,vunet} utilized GAN~\cite{goodfellow2014generative} or VAE~\cite{vae} to this task because these models are effective to infer the image given a conditional input from the learned distribution of the training data. However, due to the large variation of the poses and appearances, the quality of their generated images are still not satisfactory. 
	
	We start with the perspective that the images of all the possible poses and views of a certain person constitute a manifold, which is also suggested in previous works \cite{posemanifold2,posemanifold1}. In this perspective, pose transfer is about going from point $\mathbf{p}_{x}$ on the manifold to another point $\mathbf{p}_{y}$, both indexed by their respective poses. Therefore, the challenges mentioned above is attributed to climbing the complex structure of the manifold from $\mathbf{p}_{x}$ to $\mathbf{p}_{y}$ on the global level. However, climbing the structure manifold becomes simpler on a local perspective. If we restrict the variation of pose or appearance to a small range, pose transfer could be much simpler. For example, it is hard to transfer pose from sitting to standing, but much simpler to only raise a straight arm to a different angle. 
	
	This insight motivates us to take a \emph{progressive} pose transfer scheme. In contrast to the one-step transfer scheme adopted in some previous works~\cite{vunet,DeformableGAN}, we propose to transfer the pose of the condition image by transferring through a sequence of internal network blocks before reaching the target. The whole transfer process is carried out by a sequence of \emph{Pose-Attentional Transfer Blocks} (PATBs), where each block is responsible for a sub-step transfer. This scheme allows each transfer block to perform its transfer on the local manifold with simpler structure, therefore avoiding capturing the complex structure of a global transfer.

	In our conference version~\cite{PoseTransfer}, each PATB performs the transfer in a \emph{pose-attentional} manner.
	It takes as input the representation of both the image and the poses. Inside the block, we create an attention mechanism that infers the regions of interest based on the human pose.
	As a person's image and pose are registered, this attention mechanism allows better selectivity in choosing the image regions for transferring. The block outputs the updated image and pose representations, so that such blocks can be cascaded in sequence to form a PAT network (PATN), as depicted in Fig.~\ref{fig:Generator}.

{In this paper, we revise the design of PATB and devise another pose-attentional mechanism to preserve more appearance details. Each PATB \emph{implicitly} considers the relationship between the condition pose and the target pose, and the predicted attention mask is directly multiplied to the appearance feature in an element-wise manner. Therefore, PATB lacks the ability to directly ``move'' a feature patch to somewhere else in the image. Such difficulty gets more intensified for complex appearance features. In light of this, we design a new pose-attentional mechanism to \emph{explicitly} build the pose relationship between the condition pose and target pose, and then follow this relationship to put sampled appearance features to places indicated by the target pose. We name such explicitly calculated pose relationship as \emph{pose alignment} and the corresponding pose-attentional mechanism as \emph{aligned pose-attentional mechanism}. When equipped with such mechanism, the resulted \emph{Aligned Pose-attentional Transfer Block} (APATB) has better ability to maintain appearance features. The cascade of several APATBs eventually stimulates the generator to produce images with high quality and appearance consistency with the condition image.}

To summarize, the contributions of our paper are three fold:

	
\begin{enumerate}
	\item We propose a progressive pose attention transfer network to address the challenging task of pose transfer, which is neat in design and superior in efficacy.
	\item {The proposed network embeds novel cascaded Pose-Attentional Transfer Blocks (PATBs) or Aligned Pose-attentional Transfer Blocks (APATBs) that smoothly transfer the appearance features with the pose-attentional mechanism.}
	\item {We demonstrate that our method is able to produce augmented training data to improve the performance of  person re-identification, when the original training data is not sufficient.}
\end{enumerate}

\section{Related work}

{In recent years, a rich body of literature concerning Generative Adversarial Networks (GAN)~\cite{goodfellow2014generative} has been published. GANs, consisting of a generator and a discriminator, are able to generate sharp images~\cite{goodfellow2014generative,cgan,radford2015dcgan,perceptualloss,SRGAN,ACGAN,pix2pix2017-cvpr,CycleGANICCV,ma2018exemplar,yu2018free} through adversarial learning. The conditional generative adversarial network (CGAN)~\cite{cgan} is proposed to generate images under conditional constraints, \emph{e.g.}, specific categories or attributes. CGANs achieve remarkable success in pixel-wise aligned image generation applications. Isola \emph{et al.}~\cite{pix2pix2017-cvpr} leveraged CGAN to image-to-image translation tasks such as day-to-night and sketch-to-image. However, common CGANs are not suitable for the pose transfer task due to the large deformation between the condition and target pose. In the following, we first review several works that centers on the pose transfer task by using different pose representations. Then we present a brief introduction to several other related works.}


\vspace{1ex}\noindent\textbf{Pose transfer based on keypoints.}~Ma \emph{et al.} \cite{poseguided} presented a two-stage model to generate person images but their coarse-to-fine strategy requires relatively large computational budget and complicated training procedures. Similarly, Zhao \emph{et al.} \cite{multiview} adopted a coarse-to-fine method for generating multi-view cloth images from a single view cloth image. Ma \emph{et al.} \cite{Disentangled} further improved their previous work~\cite{poseguided} by disentangling and encoding foreground, background and pose of the input image into embedding features then decodes them back to an image. Though the controllability of the generation process is improved, the quality of their generated images degrade. Likewise, Essner \emph{et al.} \cite{vunet} exploited to combine VAE \cite{vae} and U-Net \cite{pix2pix2017-cvpr} to disentangle appearance and pose. However, appearance features are difficult to be represented by a latent code with fixed length, giving rise to several appearance misalignments. Recently, many works~\cite{DeformableGAN,dong2018soft} attempt to warp the condition image to align with the target pose. Siarohin \emph{et al.} \cite{DeformableGAN} introduced deformable skip connections that require extensive affine transformation computation to deal with pixel-to-pixel misalignment caused by pose differences. And their method is fragile to inaccurate pose disjoints, resulting in poor performance for some rare poses. Balakrishnan \emph{et al.} \cite{unseenpose} presented a GAN network that decomposes the person image generation task into foreground and background generation, and then combines to form the final image. Pumarola \emph{et al.} \cite{unsupervisedposetransfer} adopted a bidirectional strategy to generate person images in a fully unsupervised manner that may induce some geometric errors as pointed out in their paper.

\vspace{1ex}\noindent\textbf{Pose transfer based on other pose representations.}~Human body keypoint is a commonly used pose representation for the person image generation task, \emph{e.g.}, pose transfer and virtual try on, since it can be obtained robustly and cheaply by keypoint detectors. Recently, several other pose representations, such as human parsing map~\cite{dong2018soft}, 3D human model~\cite{Intrinsic} and DensePose~\cite{neverova2018dense}, are also exploited by the community. Human parsing map provides fine-grained semantic information for a person, such as clothes and body parts. However, the methods using parsing map highly rely on the accuracy of the estimated parsing map. Inaccurate parsing map could lead to obviously implausible results. Besides, it requires some complex pre-processing procedures to group multiple semantically similar categories into a single one, which is sometimes hard to handle due to the varieties of clothes styles. Dong \emph{et al.}~\cite{dong2018soft} utilized a soft-gated warping block to warp the condition image through a transformation grid generated from the parsing maps of the condition and target images. 
{Song \emph{et al.}~\cite{song2019unsupervised} proposed an unsupervised framework which decomposes the person image generation task into two subtasks, namely semantic parsing transformation and appearance generation.}
\edit{Li \emph{et al.}~\cite{Intrinsic} introduced a typical example that leverages 3D human model prior to predict dense and intrinsic appearance flow between two poses that guides the transfer of pixels. However, 3D human model is even harder to construct and generalize to complex poses and body shapes.}
DensePose maps the pixels in the images to a common surface-based coordinate system and contains abundant information of depth and body part segmentation. However, it is challenging to estimate the DensePose precisely as pointed in \cite{han2019clothflow}, limiting the application of dense pose in real-world scenarios. Neverova \emph{et al.} \cite{neverova2018dense} adopted \emph{DensePose} \cite{guler2018densepose} as its pose representation in order to present more texture details in their generated images. Grigorev \emph{et al.}~\cite{grigorev2018coordinate} introduced a coordinate-based texture inpainting method to transfer the appearance from the condition view to the target view based on the DensePose guidance.
Compared to these pose representations, keypoint is much cheaper, more accurate and flexible and thereupon serves as our only pose representation.

\edit{\vspace{1ex}\noindent\textbf{Video pose transfer.}~This task aims to transfer the appearance of a given person to targets from videos. Several works~\cite{everybody,vid2vid,fewshotvid2vid,Vid2Game} regard this task as an image-to-image translation procedure and explores available image translation frameworks~\cite{pix2pix2017-cvpr,pix2pixhd,SPADE}. These works usually include additional strategies to ensure temporal consistency of adjacent frames. Some other works~\cite{animating,firstorder} decompose appearance and motion information using a self-supervised strategy and then yield dense optical-flow to perform pixel warping, which avoids the use of strong person priors. In this paper, we mainly focus on image-level pose transfer, however, we believe our approach is not restricted and can be extended to video pose transfer by imposing a temporal consistency constraint.}
	
\vspace{1ex}\noindent\textbf{Virtual try-on.}~Virtual try-on~\cite{viton,cpviton,humanappearancetransfer,garmenttransfer,dong2019towards} is an application that has high relevance to the pose transfer task. It aims at changing the clothes of a person by warping the given clothes to fit for the body topology of the given person. Lassner \emph{et al.}~\cite{LassnerPG17} presented a model combining VAE \cite{vae} and GAN together to generate images of a person with different clothes, given the 3D model of the person. Han \emph{et al.}~\cite{viton} proposed a coarse-to-fine framework which first generates the warped clothing image through shape context matching and then outputs the person image with a refinement network. Han \emph{et al.}~\cite{han2019clothflow} further introduced a flow-based model to offer the visual correspondence between the condition image and target image guided by their parsing maps. Wu \emph{et al.}~\cite{wu2019m2e} proposed a new virtual try-on network named M2E-TON to transfer the clothes from a given model image directly without the need of clean clothes images. 
	

\vspace{1ex}\noindent\textbf{Person image generation for data augmentation.} It draws increasing attention from the community to apply the generative models to produce images for data augmentation due to the expensive cost of collecting and labeling real samples. Consequently, many works focused on using generated persom images to boost the performance of person recognition tasks like pedestrain detection and person re-identification \emph{etc.} Wu~\emph{et al.}~\cite{wupmc} introduced a multi-modal cascaded generative adversarial network to generate pedestrian images for pedestrian detection. Liu~\emph{et al.}~\cite{liu2019adaptive} proposed a adaptive transfer network, which aims at transferring a persons image into a different style, to alleviate the challenge of cross-domain person Re-Identification. Similarly, Wang \emph{et al.}~\cite{wang2019rgb} attempted to improve the RGB-Infrared cross-domain re-identification by generating infrared images from RGB images with AlignGAN.
	
\vspace{1ex}\noindent\textbf{Non-local attention.}~In \cite{SAGAN}, Zhang~\emph{et. al.} demonstrated that using self-attention mechanism in GAN models to explore the long-range dependencies in images is able to improve the generated image quality. Our improved form of pose attention is quite related to it when speaking of the goal to capture global pairwise similarities. A major difference is that our input is multi-source while SAGAN feeds a single source of feature map for its self-attention block.
    


\section{Model}
	\begin{figure}
		\centering
		\includegraphics[width=0.5\textwidth]{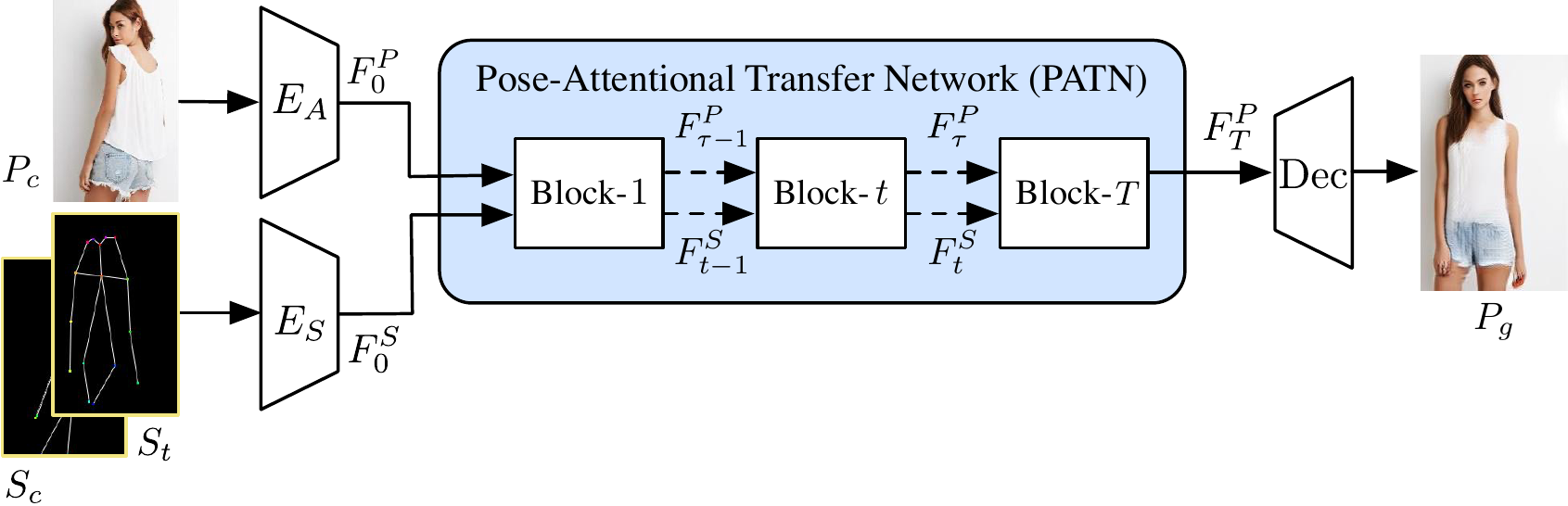}
		\caption{Generator architecture of the proposed method. $E_A$ and $E_S$ are the appearance encoder and shape encoder, respectively. $\mathrm{Dec}$ represents the decoder. \label{fig:Generator}}
	\end{figure}

\begin{figure*}[htbp]
	\centering
	\includegraphics[width=\textwidth]{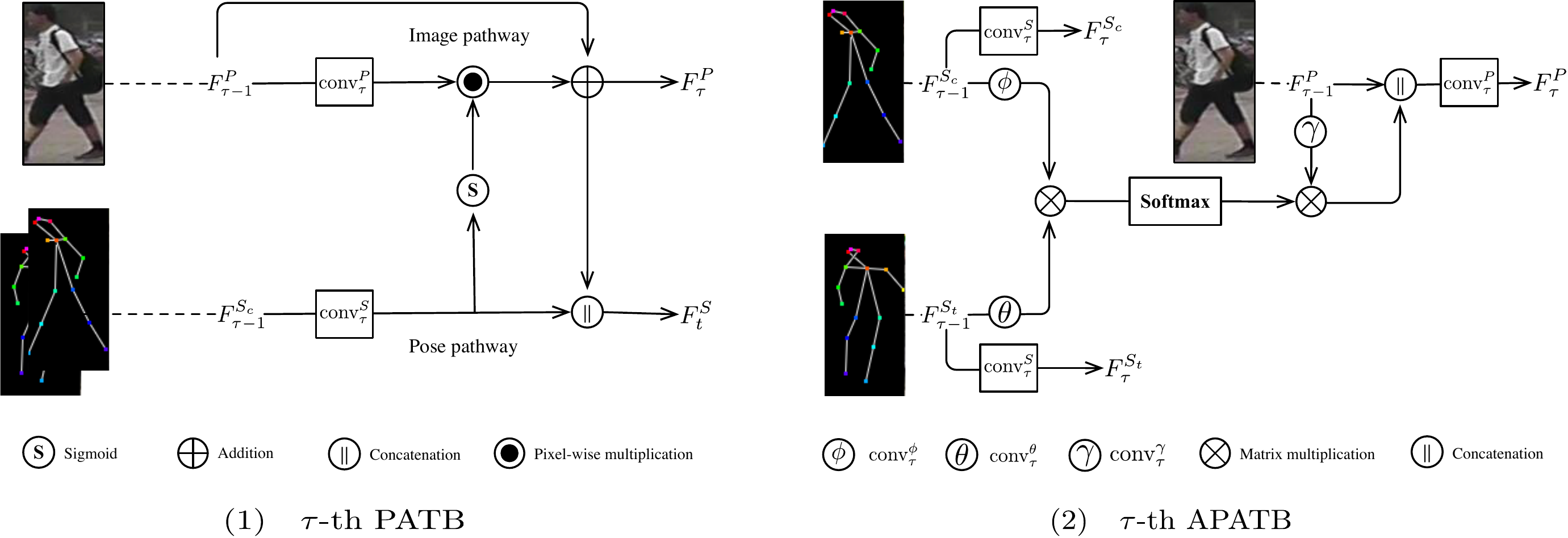}
	\caption{Detailed architectures of the two different versions of the core transfer block in our generator, namely the PATB in the left and the APATB in the right.\label{fig:PATB}}
\end{figure*}

\label{sec:network_architecture}
We begin with a few notations. $\left\{ P_{i}^{j}\right\}_{i=1\dots N_{j}}^{j=1\dots M}$
denotes the set of person images in a dataset, where $j$ is person index, $i$ is the image index of person $j$. $M$ is the number of persons, $N_j$ is the number of images of person $j$. 
$S_{i}^{j}$ is the corresponding keypoint-based representation of $P_{i}^{j}$, which consists of a 18-channel heat map that encodes the locations of 18 joints
of a human body. 
We adopt the Human Pose Estimator (HPE) \cite{HPE} used by \cite{poseguided,Disentangled,DeformableGAN} to estimate the 18 joints for fair comparison. During training, the model requires condition and target image $(P_c, P_t)$ and their corresponding condition and target pose heat map $(S_c,S_t)$. The generator outputs a person image, which is challenged by the discriminators for its realness.

Fig.~\ref{fig:Generator} shows the architecture of the generator, which requires the inputs of the condition image $P_c$, the condition pose $S_c$ and the target pose $S_t$. The main aim of the generator is to transfer the pose of the person in the condition image $P_c$ from condition pose $S_c$ to target pose $S_t$, thus generates realistic-looking person image $P_g$. In general, our generator consists of two encoders, a pose transfer network, and a decoder, which are detailed in the following.

\subsection{Encoders}
We employ two encoders in our generator: one encoder, called \emph{Appearance Encoder}, denoted as $E_A$, is used to encode the condition image to an \emph{image code}; the other encoder, named \emph{Shape Encoder}, denoted as $E_S$, is used to map the condition pose $S_c$ and the target one $S_t$ to corresponding \emph{pose codes}. In our setting, each encoder is comprised of one $7 \times 7$ convolution and two $3 \times 3$ convolutions with stride 2. We add a BatchNorm layer and a ReLU activation function after each convolution layer. Such a lighted-weighted network encodes valid features and saves the computation overhead for the following pose transfer process.




\subsection{Pose Transfer Network}
The core of our generator is the Pose Transfer Network. This network takes the image code and pose code as inputs and then learns to transfer the person image under condition pose $S_c$ to the target pose $S_t$. At a basic level, pose transfer is about moving patches from
the locations induced by the condition pose to the locations induced by the target pose. In this sense, the pose guides the transfer process by hinting \emph{where to sample condition patches} and \emph{where to put target patches}. However, large pose variations brings great challenge for this \textit{sample-and-put} operation. To address the issue, we design a progressive pose transfer network which consists of a series of pose transfer blocks. \edit{Besides, the network can also benefit from large-scale training data to infer unobserved parts or pixels, though they are not explicitly handled.}

The design of the pose transfer block is crucial for the Pose Transfer Network. In our conference version, we propose a \textbf{Pose Attentional Transfer Network}~(PATN) with multiple \textbf{Pose-Attentional Transfer Blocks}~(PATB). {In this paper, we present another type of PATB called \textbf{Aligned Pose-Attentional Transfer Block}~(APATB), which utilizes an aligned pose-attentional mechanism to establish the alignment between the condition pose and the target pose.} For convenience, in the next sections, we call the PATN whose internal blocks are changed to APATBs as the \textbf{Aligned Pose-Attentional Transfer Network}~(APATN).

In the following, we first present a brief review of the PATN and the PATB. Then we state the difference between PATB and APATB. Finally, we demonstrate the design of the APATB.

\subsubsection{Pose-Attentional Transfer network}

Pose-Attentional Transfer Network (PATN) consists of several cascaded Pose-Attentional Transfer Blocks (PATBs). For PATN, $S_c$ and $S_t$ are concatenated together and then fed to $E_S$. Therefore, the initial pose code is the joint combination of the two poses, denoted as $F_{0}^{S}$. From the input side, PATN also receives an initial image code $F_{0}^{P}$, encoded by $E_A$. On the whole,
PATN progressively updates these two codes through a sequence of PATBs. At the output, the final image code $F_T^P$ are taken to decode the output image, while the final pose code $F_T^S$ is discarded.

All PATBs have identical structure. A PATB carries out one
step of update. Consider the $\tau$-th block, whose inputs are $F_{\tau-1}^{P}$
and $F_{\tau-1}^{S}$. As depicted in Fig.~\ref{fig:PATB} (1), the block comprises two pathways, called \emph{image pathway}
and \emph{pose pathway} respectively. With interactions, the two
pathways update $F_{\tau-1}^{P}$ and $F_{\tau-1}^{S}$ to $F_{\tau}^{P}$
and $F_{\tau}^{S}$, respectively. 
The interaction is mainly conducted by an attentional mechanism. The pose pathway produces element-wise pose attention mask $M_{\tau}$ for the image pathway by feeding the pose code $F_{\tau-1}^{S}$ through a typical residual block $\text{conv}^{S}_{\tau}$~\cite{resnet} and a Sigmoid layer $\sigma$ that maps the outputs of the previous residual block to value interval of $[0, 1]$. Such procedure can be briefly described as:
\begin{equation}
	M_{\tau} = \sigma(\text{conv}^{S}_{\tau}(F_{\tau-1}^{S}))
\end{equation}
 The image code goes through another residual block $\text{conv}^{P}_{\tau}$ and the resulted feature map is multiplied with the attention mask in an element-wise manner:
 \begin{equation}
 \hat{F^P_{\tau}} = M_{\tau}\odot\text{conv}^{P}_{\tau}\left(F_{\tau-1}^{P}\right).
\end{equation} $\hat{F^P_{\tau}}$ represents the attended result of the previous image code and is then added by $F^P_{\tau-1}$ , constituting a residual connection. Having such residual connection helps reserve important features and eases the training~\cite{resnet}, particularly when there are many PATBs. Hence, the image code $F_{\tau}^{P}$ of the $\tau$-th PATB is:
\begin{equation}
F_{\tau}^{P}=\hat{F^P_{\tau}}+F_{\tau-1}^{P},
\end{equation} where $\odot$ denotes element-wise multiplication.
As the image code gets updated through the PATBs, the pose code should also be updated to synchronize the change, \emph{i.e.} 
where to sample and put patches given the new image code. Therefore,
the pose code update should incorporate the new image code, which is conducted by:
\begin{equation}
F_{\tau}^{S}=\mathrm{conv}^{S}_{\tau}\left(F_{\tau-1}^{S}\right)\Vert F_{\tau}^{P},
\end{equation} where $\Vert$ denotes the concatenation of two maps along depth axis. 


\subsubsection{Aligned Pose-Attentional Transfer Network}
\label{PMTN}
\vspace{1ex}\noindent\textbf{Problem analysis.}~APATB is the core of the Aligned Pose-Attentional Transfer Network, just as PATB is the core to PATN. With thorough investigation of the PATB, we found one critical issue that may limit its performance. PATB feeds the combined pose features of the condition and target pose to a residual block to produce the attention mask directly. However, it ignores to explicitly leverage the spatial alignment between the condition and target pose, which is helpful to smoothly move the features to correct positions. Hence, PATB inherently has difficulty in effectively conducting the \textit{sample-and-put} operation, leading to sub-optimal performance. 
As a contrast, APATB explicitly calculates the alignment between the condition pose and target pose, and then aggregates the features of the image code according to this pose alignment. 
This new attention mechanism enables more direct and smooth feature transfer, thus it largely helps preserve appearance details. In the following, we depict the structure of APATB with three sub-sections: \emph{pose alignment calculation}, \emph{image code update} and \emph{pose code update}.



\label{sec:PMTN}


\vspace{1ex}\noindent\textbf{Pose Alignment calculation.}~Explicitly calculating the pose alignment requires the independence of the condition pose code and target pose code, which differs from PATB. We denote the initial condition pose code and target pose code as $F_0^{S_c}$ and $F_0^{S_t}$, respectively. 
The structure of APATB is depicted in Fig.~\ref{fig:PATB} (2). 
Considering the pose codes are actually spatial maps in the feature space, an intuitive way to calculate pose alignment is to gather the similarities between all the pixels in the condition pose code with all the pixels of the target pose code, which could be simply implemented as a matrix inner production.
Specifically, we first feed the condition pose code $F_{\tau-1}^{S_c}$ and the target pose code $F_{\tau-1}^{S_t}$ to a $1\times1$ convolution layer to reduce the channel dimension to a reasonable number as $C$. This helps save computation and GPU-memory occupation while accelerating the training and inference process. Then we resize the pose codes with the shape of $H \times W \times C$ into the shape of $(HW) \times C$, where $H$, $W$, $C$ denote the height, the width, and the channel dimension, respectively. 
Afterwards, we calculate the pose alignment $A_{\tau}$ between these two pose codes in a matrix inner production manner, which can be formulated as:
\begin{equation}
A_{\tau} = \mathrm{conv}^{\theta}_{\tau}(F_{\tau-1}^{S_t}) \times \mathrm{conv}^{\phi}_{\tau}(F_{\tau-1}^{S_c})^T
\end{equation}
where $\mathrm{conv}^{\theta}_{\tau}(\cdot)$ and $\mathrm{conv}^{\phi}_{\tau}(\cdot)$ denote the $1\times 1$ convolution layer, $\times$ represents the matrix inner production. 
The pose alignment $A_{\tau}$ is in the shape of $(HW) \times (HW)$, indicating the similarity of all the pairs of the spatial positions. We apply a Softmax function to $A_{\tau}$ so that sum of each row of $A_{\tau}$ is equal to 1. In this way, for each spatial location $L_{x,y}$ in the pose code, where $0\leq x < H, 0\leq y < W$, $A_{\tau}$ stores the attentional weights of all spatial locations to $L_{x,y}$. Thereafter, the calculated pose alignment matrix is functionally equivalent to the attention mask in PATB.

\vspace{1ex}\noindent\textbf{Image code update.}~After requiring the element-wise pose alignment matrix, the next step is to transfer image features according to the guidance of pose alignment. Such step can be realized by computing the response at a position as a weighted sum of the features at all positions in the image code, via another matrix inner production operation between $A_{\tau}$ and $F_{\tau-1}^{P}$. Similar to the channel dimension reduction strategy applied to the pose codes, we also leverage a $1\times1$ convolution, denoted as $\text{conv}_{\tau}^{\gamma}$, to reduce the channel number of the image code $F_{\tau-1}$. Then, such matrix inner production can be written as:
\begin{equation}
\hat{F_{\tau}^{P}}=A_{\tau} \times \text{conv}_{\tau}^{\gamma}(F_{\tau-1}^{P}),
\end{equation}
where $\hat{F_{\tau}^{P}}$ refers to a temporary image code generated via this aligned pose-attentional mechanism. 
At early stages of the training phase, the calculated pose alignment $A_{\tau}$ may be inaccurate, which could possibly lead to uninformative $\hat{F_{\tau}^{P}}$ and training divergence. To avoid this problem, we resort to borrowing the cues from the previous image code to reduce this information loss, done through a concatenation. Then, the image code is updated by:
\begin{equation}
F_{\tau}^{P}=\text{conv}_{\tau}^{P}\left(\hat{F_{\tau}^{P}} \Vert F_{\tau-1}^{P}\right).
\end{equation}
Note we can also use addition between $F_{\tau-1}^{P}$ and $\hat{F_{\tau}^{P}}$ to fuse them together. However, we experimentally found concatenation leads to better performance. We hypothesize that concatenation is able to propagate more detailed information and provide more flexibility for the later residual block to opt to a better state. 





\vspace{1ex}\noindent\textbf{Pose code update.}~Since APATN adopts pose heat maps as the pose representation, the pose codes tends to be spatially sparse. Consequently, with larger input, APATB may lack sufficient receptive field to capture the topology of the human body which is beneficial to build more accurate pose alignment. To alleviate this problem, we update the pose codes by feeding them to a residual block inside each APATB to progressively enlarge the receptive field. The updated pose codes are therefore embedded with stronger power to express its inner pose joints. 
This update process is described as:
\begin{equation}
F^{S_c}_{\tau} = \text{conv}_{\tau}^S(F^{S_c}_{\tau-1}), \ 
F^{S_t}_{\tau} = \text{conv}_{\tau}^S(F^{S_t}_{\tau-1}).
\end{equation}
Note we use the same residual block to update the condition pose and target pose separately because it helps to constrain the value range of the two pose codes in the same level, which potentially makes it easier to calculate the similarities of the two pose codes and also saves parameters.

\subsection{Decoder}
The updates in the pose transfer network result in the final image code $F_{T}^{P}$. The decoder recovers the image resolution and generates the transferred image $P_{g}$ from $F_{T}^{P}$ via several upsampling operations. 
In our setting, the decoder is adopted as a similar architecture to our encoder. Concretely, we adopt two bilinear upsamling layers and a $7\times7$ convolution layer followed by a tanh activation function. We add a residual block after each upsampling layer to enhance the capacity and smooth the feature representation.

\subsection{Discriminators}

We design two discriminators called \emph{appearance discriminator} $D_{A}$ and \emph{shape discriminator} $D_{S}$, to judge how likely $P_{g}$ contains the same person in $P_{c}$ (\emph{appearance consistency}) and how well $P_{g}$ align with the target pose $S_{t}$ (\emph{shape consistency}).

The two discriminators
have similar structures, where $P_{g}$ is concatenated with either
$P_{c}$ or $S_{t}$ along the depth axis, before being fed into a
CNN (convolutional neural network) for judgment. Their outputs are respectively $R^{A}$ and $R^{S}$,
\emph{i.e.} the appearance consistency score and the shape consistency
score. The scores are the probabilities output by the Sigmoid layer
of a CNN. The final score $R$ is the production of the two scores: $R=R^{A}R^{S}$.

As the training proceeds, we observe that a discriminator with low capacity becomes insufficient to differentiate real and fake data. 
Therefore, we build the discriminators by adding three residual blocks after two down-sampling convolutions to enhance their capability.

\subsection{Loss function}

\label{lossfunction}
	The full loss function is denoted as:
	\begin{equation}
	\mathcal{L}_{full}=\arg\underset{G}{\min}\, \underset{D}{\max}\, \alpha \mathcal{L}_{GAN} + \lambda_1 \mathcal{L}_{L1}+\lambda_2 \mathcal{L}_{perL1},,
	\end{equation}
	where $\mathcal{L}_{GAN}$ denotes the adversarial loss, $\mathcal{L}_{L1}$ denotes the pixel-wise $L_1$ loss computed between the generated and target image and $\mathcal{L}_{L1}=\left \| P_{g}-P_t \right \|_1$ and $\mathcal{L}_{perL1}$ represents the perceptual loss widely used in the tasks of super-resolution \cite{SRGAN}, style transfer \cite{perceptualloss}, \emph{etc}. Note \edit{$\alpha, \lambda_1, \lambda_2$} represents the weights of $\mathcal{L}_{GAN}$, $\mathcal{L}_{L1}$ and $\mathcal{L}_{perL1}$, respectively. We empirically find that perceptual loss is helpful to reduce pose distortion and make the produced images look more natural and smooth. This type of loss is also used in other pose transfer methods~\cite{vunet}. In formula, 
	\begin{small}
		\begin{equation}
		\mathcal{L}_{perL1}=\frac{1}{W_{\rho}H_{\rho }C_{\rho}} \sum_{x=1}^{W_{\rho }}\sum_{y=1}^{H_{\rho }}\sum_{z=1}^{C_{\rho }}\left \| \phi _{\rho }(P_{g})_{x,y,z}-\phi _{\rho }(P_t)_{x,y,z} \right \|_1,
		\end{equation}
	\end{small}where $\phi _{\rho }$ are the outputs of a layer, indexed by $\rho$, from the VGG-19 model \cite{vgg} pre-trained on ImageNet \cite{ImageNet}, and $W_{\rho}$,$H_{\rho }$,$C_{\rho}$ are the spatial width, height and depth of $\phi _{\rho }$, respectively. We found $\rho = \text{Conv1}\_ 2$ leads to the best results in our experiments. 
	
	The adversarial loss is derived from $D_{A}$ and $D_{S}$:
	
	\begin{small}
		\begin{equation}
		\begin{split}
		\mathcal{L}_{GAN}=&\mathbb{E}[\log [D_{A}(P_c,P_t) + (1-D_{A}(P_c,P_g)]]\\ 
		+&\mathbb{E}[\log [D_{S}(S_t,P_t) + (1-D_{S}(S_t,P_g))]].
		\end{split}
		\end{equation}
	\end{small}Note that $(P_c, P_t)$, $(S_c, S_t)$ and $P_g$ are in the distributions of real person images, real person poses and fake person images, respectively. 
	
	\vspace{1ex}\noindent\textbf{Implementation details.}
	The implementation is built upon the popular Pytorch framework.
	We adopt Adam optimizer \cite{Kingma2014-Adam} to train the proposed model for around 90k iterations with $\beta_1=0.5,\beta_2=0.999$. Learning rate is initially set to $2\times 10^{-4}$, and linearly decay to 0 after 60k iterations. APATN has 5 APATBs while PATN has 9 PATBs for both datasets. \edit{($\alpha,\lambda_1$, $\lambda_2$)} is set to (5, 1, 1). We use instance normalization~\cite{instancenorm} as the normalization method in both the generator and the discriminators. Dropout \cite{dropout} is only applied in the PATBs and APATBs with the dropout rate set to 0.5. Leaky ReLU \cite{LeakyReLU} is applied after every convolution or normalization layers in the discriminators, and its negative slope coefficient is set to 0.2. Our code is available at \url{https://github.com/tengteng95/Pose-Transfer}.

\section{Experiments}
In this section, we conduct extensive experiments to verify the efficacy and efficiency of the proposed network. The experiments not only show the superiority of our network but also verify its design rationalities in both objective quantitative scores and subjective visual realness.



\subsection{Datasets}
\vspace{1ex}\noindent\textbf{Market-1501~\cite{Market1501}} is a challenging person re-identification dataset whose images are in low-resolution ($128\times64$) and vary enormously in the pose, viewpoint, background and illumination. We use this dataset to show the performance of our method on the pose transfer task and the ability to generate images to augment the training set for person ReID. We adopt HPE \cite{HPE} as pose joints detector and filter out images where no human body is detected. Consequently, we collect 263,632 training pairs and 12,000 testing pairs for Market-1501.

\vspace{1ex}\noindent\textbf{DeepFashion~\cite{DeepFashion}} is an \emph{In-shop Clothes Retrieval Benchmark} whose images are in high-resolution ($256\times256$) and of clean backgrounds. In this case, we use this dataset to show our method is superior to PATN in preserving local details. Likewise, we also adopt HPE \cite{HPE} as pose joints detector and filter out non-human images. Consequently we acquire 101,966 pairs for training and 8,570 pairs for testing. It's worth noting that the person identities of the training set do not overlap with those of the testing set for better evaluating the model's generalization ability.



\subsection{Metrics}
It remains an open problem to effectively evaluate the appearance and shape consistency of the generated images. 
Ma \emph{et al.} \cite{poseguided} used Structure Similarity (SSIM) \cite{SSIM} and Inception score (IS) \cite{improvedtrainingGAN} as their evaluation metrics, then introduced their masked versions, \textit{i.e.}, mask-SSIM and mask-IS, to reduce the background influence by masking it out. Siarohin \emph{et al.} \cite{DeformableGAN} further introduced Detection Score (DS) to measure whether a detector \cite{SSD} can correctly detect the person in the image. We argue that all the metrics mentioned above cannot explicitly quantify the shape consistency. More specifically: SSIM relies on global covariance and means of the images to assess the structure similarity, which is inadequate for measuring shape consistency; IS and DS use image classifier and object detector to assess generated image quality, which are unrelated to shape consistency. 

As all the metrics mentioned above are either insufficient or unable to quantify the shape consistency of the generated images,  we hereby introduce a new metric as a complement to explicitly assess the shape consistency. \edit{Specifically, we use 18 pose joints, which are obtained from the Human Pose Estimator (HPE)~\cite{HPE}, to roughly represent person shape. If the pose of the generated person aligns well to the given target pose, we assume the generated person is consistent in shape. In pose estimation task, one common measure to assess the joint distances between the predicted pose and ground-truth pose is the PCKh measure}\footnote{PCKh is the slightly modified version of Percentage of Correct Keypoints (PCK) \cite{PCK}} \cite{PCKh}. In this paper, we also adopt this measurement to assess the shape consistency between the poses of generated persons and the given target poses. According to the protocol of \cite{PCKh}, PCKh score is the percentage of the keypoints pairs whose offsets are below the half size of the head segment. The head is estimated by the bounding box that tightly covers a set of keypoints related to head.

Besides, we also test the running speed of the models and report their total number of parameters in the generator. We believe faster and more light-weight models are more close to real-world applications and can be potentially employed on mobile devices.

\subsection{Comparison with previous work}
\label{sec:comparison_with_previous_work}

\subsubsection{Quantitative and qualitative comparison}

\begin{table*}[htbp]
	\centering
	\caption{Comparison with state-of-the-art on Market-1501 and DeepFashion. * denotes the results tested on our test set.}
	\resizebox{\textwidth}{!}{
		\begin{tabular}{|l|cccccc|cccc|}
			\hline
			\multirow{2}{*}{Model} & \multicolumn{6}{c|}{Market-1501}           & \multicolumn{4}{c|}{DeepFashion} \\ \cline{2-11} 
			& SSIM  & IS    & mask-SSIM & mask-IS & DS  & PCKh & SSIM      & IS       & DS    & PCKh   \\ \hline
			Ma \emph{et al.} \cite{poseguided}  & 0.253 & 3.460          & 0.792     & 3.435   & -     & -     & 0.762    & 3.090   & -   & -     \\
			Ma \emph{et al.} \cite{Disentangled}    & 0.099 & 3.483 & 0.614     & 3.491   & -     & -     & 0.614    & 3.228    & -       & -     \\
			Siarohin \emph{et al.} \cite{DeformableGAN}           & 0.290 & 3.185          & 0.805     & 3.502   & 0.72  & -     & 0.756  & 3.439    & 0.96   & -     \\  \hline
			Ma \emph{et al.} * \cite{poseguided} & 0.261 & {3.495} & 0.782     & 3.367   & 0.39  & 0.73  & 0.773    & 3.163    & 0.951   & 0.89  \\
			Siarohin \emph{et al.} * \cite{DeformableGAN}          & 0.291 & 3.230          & 0.807     & 3.502   & 0.72  & 0.94  & 0.760    & 3.362    & 0.967   & 0.94   \\ 
			Esser \emph{et al.} * \cite{vunet}          & 0.266 & 2.965 & 0.793 & 3.549 & 0.72 & 0.92 & 0.763 & {3.440} & 0.972 & 0.93 \\ 
			Tang \emph{et al.} *~\cite{XingGAN} & 0.313 & {3.506} & 0.816 & {3.872} & 0.57 & 0.93 & 0.778 & 3.476 & 0.962 & 0.95 \\ \hline
			PATN             & 0.311  & 3.323  & 0.811  & {3.773}  & {0.74}   & {0.94}  & 0.773  & 3.209 & {0.976}  & {0.96} \\
 			APATN & {0.326}	& 3.426 & {0.817}	& {3.801} & 0.66 & 0.91  & {0.780} &	3.435 & 0.971 & 0.96 \\  \hline
			Real Data  & 1.000 & 3.890  & 1.000  & 3.706  & 0.74  & 1.00   & 1.000  & 4.053    & 0.968 & 1.00        \\  \hline
	\end{tabular}}
	\label{quantitative-comparison}
\end{table*}

{To avoid the perturbation of performance caused by different data splits, we download the well-trained models of the previous keypoint-based methods~\cite{poseguided,DeformableGAN,vunet,XingGAN} from their online resources and re-evaluate their performance on our testing set. Quantitative comparisons with previous works can be found in Tab.\ref{quantitative-comparison}. \edit{Notice that~\cite{XingGAN} adopts the same training and testing split as ours while~\cite{poseguided,DeformableGAN,vunet} adopt different splits. Therefore, it is likely that our testing data overlaps with their training data but excludes our training data. It can be assumed that testing their well-trained models on our testing split shows on par with or better performance than their models trained from scratch using our training and testing split.} Under this situation, our method still outperforms~\cite{poseguided,DeformableGAN,vunet} on most metrics, and the numeric improvements are steady for both datasets. Notably, our APATN ranks the highest on the SSIM metric on both datasets, indicating the images of APATN have the best appearance consistency. Other metrics of APATN on DeepFashion draw near to PATN. On the Market-1501 dataset, APATN has significant performance improvements on SSIM, IS and mask-IS over PATN while lower DS and PCKh. The most direct reason of lower DS and PCKh is that some images generated by APATN has blurry backgrounds, which cannot be well examined by the SSD detector and the human pose detector. The problem tends to intensify on the Market-1501 dataset since the background is very complex and usually different for a single person in different images, causing confusion for APATN to align the background features correctly. \edit{Compared to a concurrent work~\cite{XingGAN} published in ECCV 2020 and built with similar ideas to PATN of our conference version~\cite{PoseTransfer}, APATN also achieves comparable results, revealing its superior performance.}} 


\begin{figure}[htbp]
	\centering
	\includegraphics[width=0.48\textwidth]{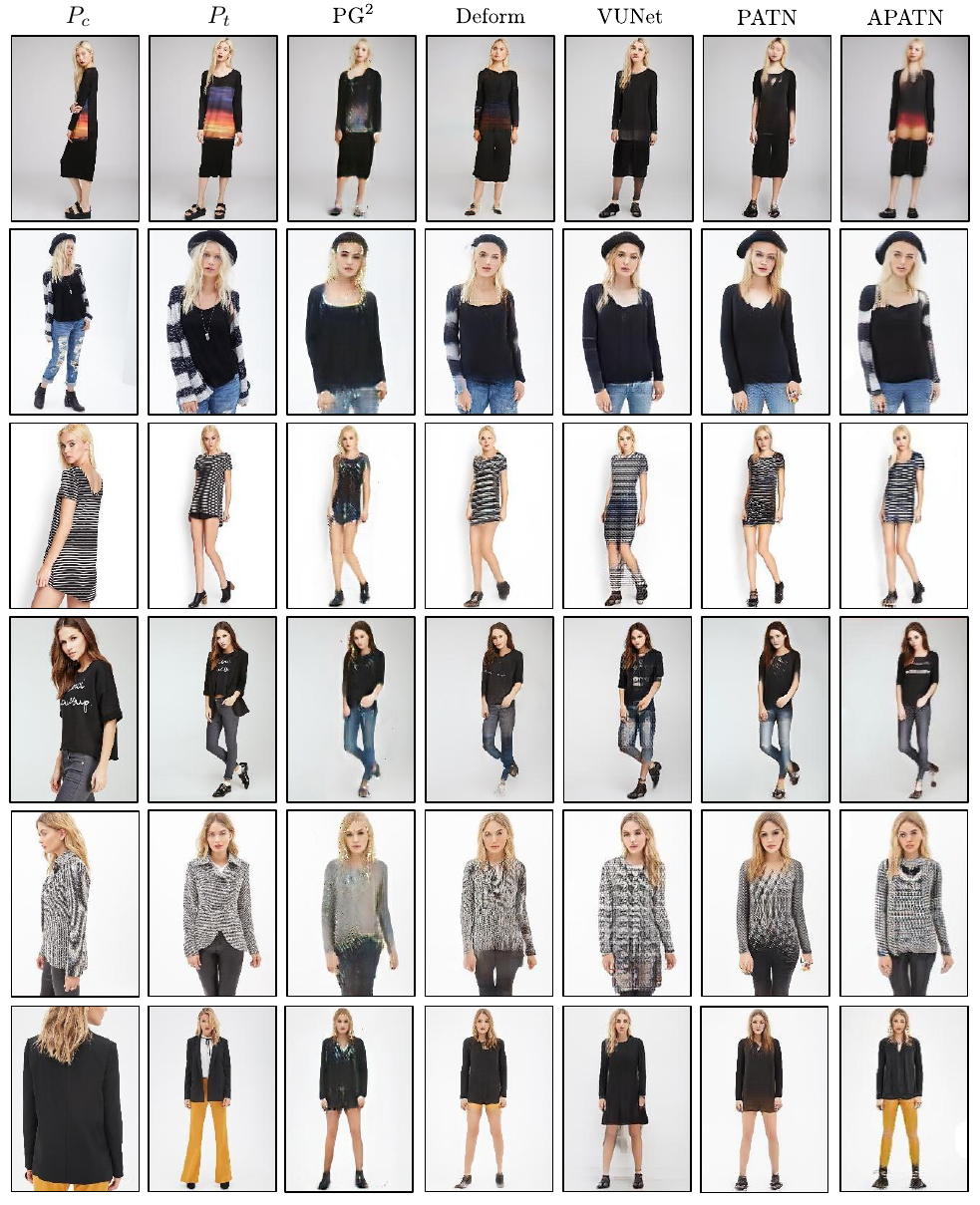}
	\caption{Qualitative comparisons on DeepFashion dataset. $\mathrm{PG^2}$, $\mathrm{VUnet}$ and $\mathrm{Deform}$ represent the results of \cite{poseguided}, \cite{vunet} and \cite{DeformableGAN}, respectively.}
	\label{fig:qualitative_comparison}
\end{figure}


{Fig.\ref{fig:qualitative_comparison} gives some typical qualitative examples with large pose changes and/or scale variance on the high-resolution dataset DeepFashion \cite{DeepFashion}. Overall, our PATN maintains the integrity of the person and exhibits natural postures. Besides, APATN demonstrate the most detailed appearance and reasonable texture. \textit{e.g.}, the pattern in the first row, the sleeve in the second row, and the stripe in the third row. Besides, APATN generates more beautiful facial details than other methods. It shows that explicitly aligning the condition pose and the target pose by the aligned pose-attentional mechanism is of great importance for keeping the appearance, especially under large pose variance.}

\begin{figure}[ht]
	\centering
	\includegraphics[width=0.48\textwidth]{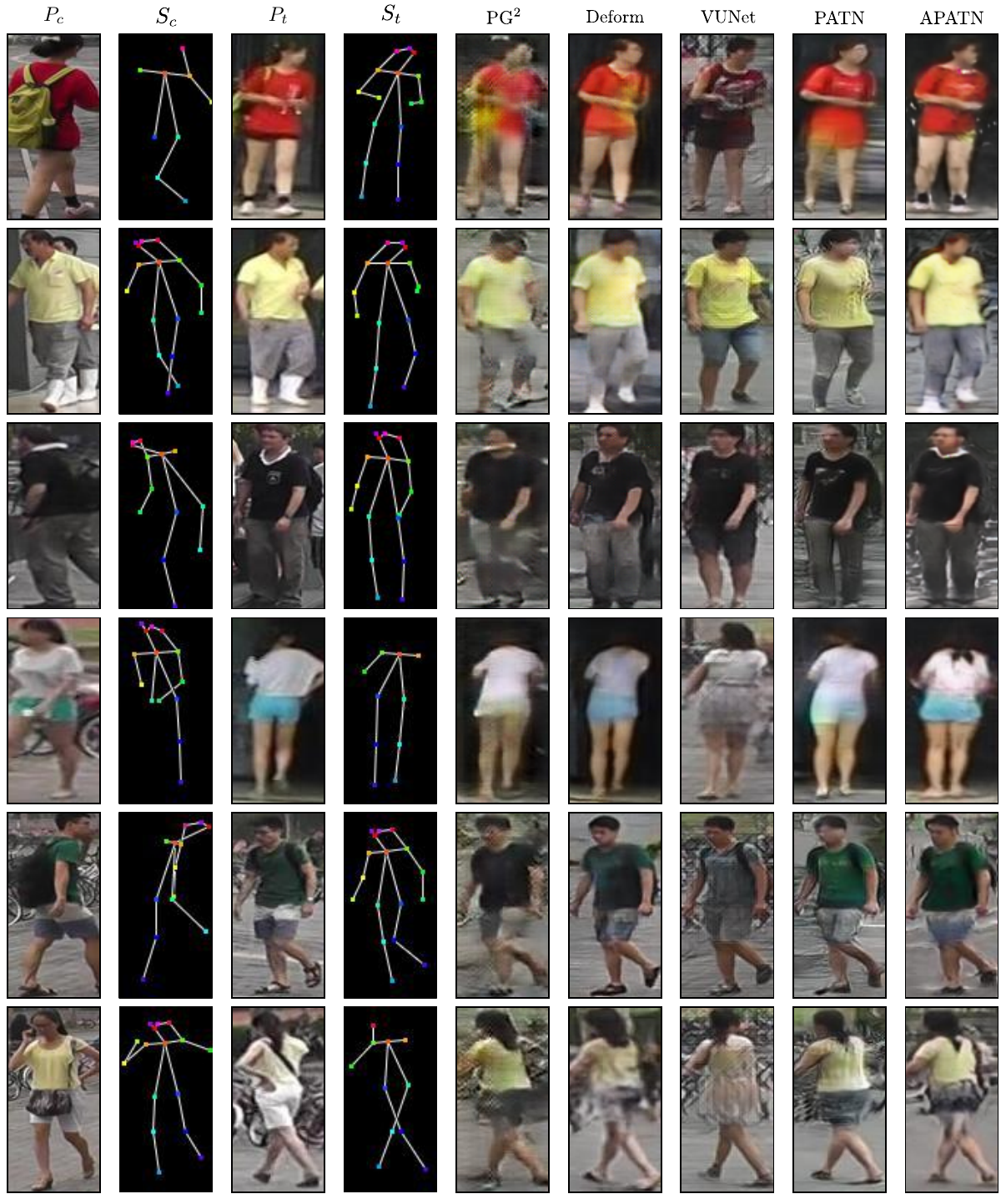}
	\caption{Qualitative comparisons on Market-1501 dataset. $\mathrm{PG^2}$, $\mathrm{VUnet}$ and $\mathrm{Deform}$ represent the results of \cite{poseguided}, \cite{vunet} and \cite{DeformableGAN}, respectively.}
	\label{fig:comparison_market}
\end{figure}


{We also evaluate the performance of our method on Market-1501, a dataset of poor image quality. Some examples are shown in Fig.~\ref{fig:comparison_market}. 
Similar to the the phenomenon observed on the DeepFashion dataset, our APATN generates images with more appearance details, such as the socks in the first row, the shoes in the second row, and the clothes around the neck in the third row. Moreover, for the case where the condition pose and/or the target pose is rare, our PATN and APATN cans still generate plausible results, while other methods leads to unsatisfying images which are blurry and/or possess incorrect appearance .}

\subsubsection{Model and computation complexity comparison}
Tab.\ref{tab:efficiency} gives the comparison for the model and computation complexity between previous keypoint guided approaches and our methods. We test these methods under one NVIDIA Titan Xp graphics card in the same workstation. Only GPU time is taken into account when generating all the testing pairs of DeepFashion to compute the speed. 

\begin{figure}
		\centering
		\includegraphics[width=0.48\textwidth]{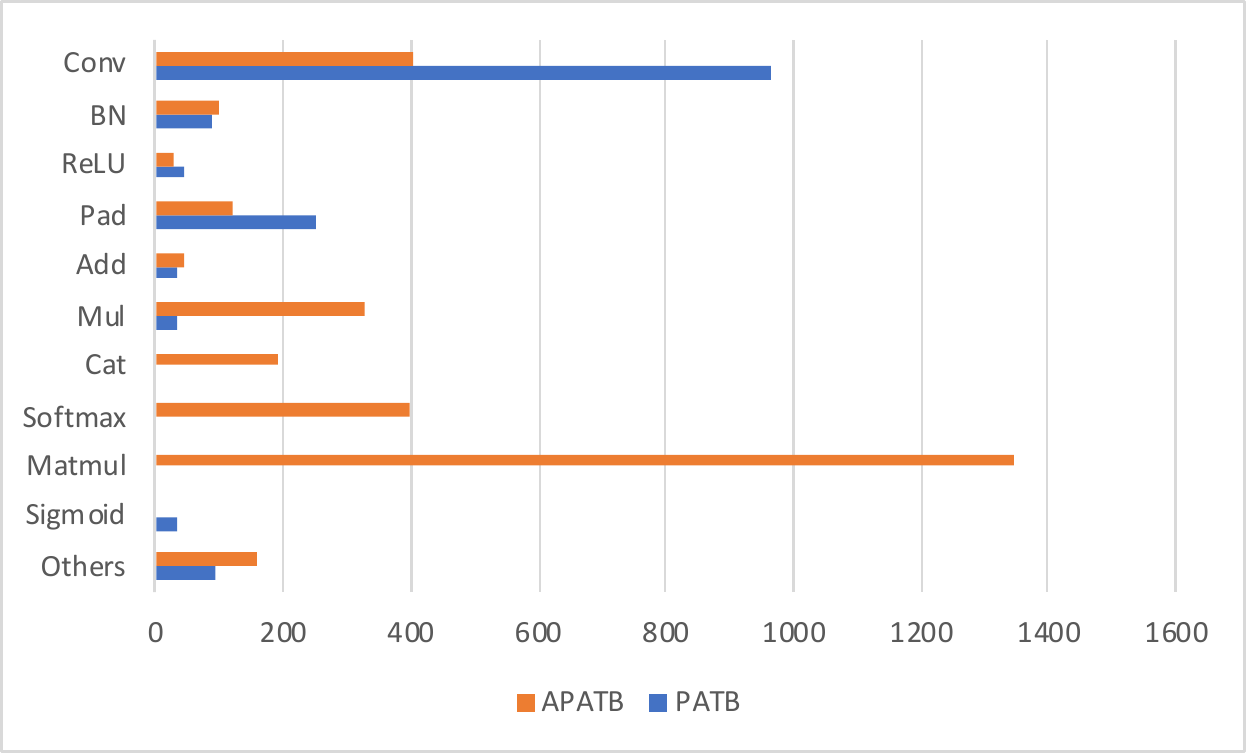}
		\caption{\edit{GPU running time comparison ($\mu$s) of different operations between a single APATB and a PATB. The collected data is averaged from 1K iterations with 1 instance per iteration. The spatial size of the input feature of both blocks is 64$\times$64 and the channel number is 256. \emph{Conv}, \emph{BN}, \emph{Mul}, \emph{Cat}, \emph{Matmul} represent convolution, batch normalization, multiplication, concatenation and matrix multiplication operations, respectively.} \label{fig:patb_time}}
\end{figure}

Notably, our networks PATN and APATN outperform other methods in terms of the number of parameters and the computation complexity, owing to the simple and neat structure of the building blocks of our network. It is worth noticing that APATN is around 1/3 of the parameter number of PATN. \edit{A single APATB has 2.76M learnable parameters and a single PATB has 4.72M learnable parameters. The main reason of the parameter reduction in a single block attributes to the convolution right after the concatenation operation: in PATB, its kernel size is set to $3\times3$ yet in APATB, it is set to $1\times1$. Therefore, though APATB has more convolutional layers, it is comparatively more parameter-saving than PATB.} Considering the superior performance of APATN under such small parameter number, we believe APATN effectively leverages its network power to perform the pose transfer task. The running speed of APATN is slightly lower than PATN because of the more computationally expensive aligned pose-attentional mechanism. \edit{A more detailed component-wise GPU running time comparison between a single APATB and PATB is given in Fig.~\ref{fig:patb_time}. The average GPU running time of the whole APATB and PATB is 3,114 $\mu$s and 1,543 $\mu$s per instance. From the figure and above statistics, the extra computation of APATB over PATB lies in the necessary operations of aligned pose-attention, such as matrix multiplication and Softmax operation. It still remains a challenging problem to bring down the computational cost of such dense non-local attentional mechanism. Though some researches achieve promising successes in semantic segmentation task by virtue of sparse sampling~\cite{ann,ccnet}, our task innately demands dense relationship calculation for preserving more details and thus poses extra challenge for our future researches.}

As for other methods, the two-stage strategy of Ma \emph{et al.} \cite{poseguided} brings a huge increase in parameters and computational burden. Although Siarohin \emph{et al.} \cite{DeformableGAN} switched to one stage model by introducing the deformable skip connections (DSCs) leading to a decreased model size, the computation-intensive body part affine transformations required by DSCs make the decrease of computation complexity only marginal. \edit{Since \cite{XingGAN} adopted similar network designs to PATN, its parameter number and running speed are relatively close to PATN.}

\begin{table}[htbp]
	\centering
	\caption{Comparison of model size and testing speed on DeepFashion dataset. ``M'' denotes millions and ``fps'' denotes Frames Per Second.
		\label{tab:efficiency} }
	\resizebox{0.4\textwidth}{!}{ 
		\begin{tabular}{|l|l|l|}
			\hline
			Method          & Params  & Speed  \\ \hline
			Ma \emph{et al.} \cite{poseguided}       & 437.09 M       &  10.36 fps   \\
			Siarohin \emph{et al.} \cite{DeformableGAN} & 82.08 M        &  17.74 fps     \\
			Esser \emph{et al.} \cite{vunet} & 139.36 M & 29.37 fps\\ 
			Tang \emph{et al.}~\cite{XingGAN} & 44.85M & 57.40 fps \\  \hline
			PATN (9 PATBs)            & 41.36 M        &  60.61 fps \\
			APATN (5 APATBs)   & 16.93 M        &  58.82 fps
			\\ \hline
	\end{tabular}}
\end{table}

\subsubsection{User study}
Human perception is more suitable to judge the realness of the generated images. To further investigate the objective realness of our generated images, we recruit 30 volunteers to give an instant judgment (real/fake) about each image within a second. Following the protocol of \cite{poseguided,DeformableGAN}, 55 real and 55 generated images are randomly selected and shuffled, the first 10 of them are used for practice then the remaining 100 images are used as the evaluation set. PATN and APATN achieve considerable performance improvements over \cite{poseguided,DeformableGAN} on all measurements, as shown in Tab.\ref{tab:user_study}, further validating that our generated images are more realistic, natural and sharp. \edit{The result of APATN is comparable to a concurrent work~\cite{XingGAN}.} It's worth noticing that our method quite excels at handling condition images of poor quality, where 70.32\% of the generated images by APATN are regarded as real by volunteers, reflected as the ``G2R'' measure in Tab.\ref{tab:user_study}. Also, for DeepFashion, APATN achieves the best score in the ``G2R'' measure.

\begin{table}[htbp]
	\centering
	\caption{User study ($\%$). \textbf{\emph{R2G}} means the percentage of real images rated as generated w.r.t. all real images. \textbf{\emph{G2R}} means the percentage of generated images rated as real w.r.t. all generated images. The results of other methods are drawn from their papers.}
	\resizebox{0.45\textwidth}{!}{ 
		\begin{tabular}{|l|cc|cc|}
			\hline
			\multirow{2}{*}{Model} & \multicolumn{2}{c|}{Market-1501}           & \multicolumn{2}{c|}{DeepFashion} \\ \cline{2-5}  & R2G & G2R & R2G & G2R\\ \hline
			Ma \emph{et al.} \cite{poseguided} & 11.2 & 5.5 & 9.2 & 14.9 \\
			Siarohin \emph{et al.} \cite{DeformableGAN} & 22.67 & 50.24 & 12.42 & 24.61 \\
			Tang \emph{et al.}~\cite{XingGAN} & 35.28 & 65.16 & 21.61 & 33.75 \\ \hline
			PATN & 32.23 & 63.47 & 19.14 & 31.78 \\ 
			APATN & 34.17 & 70.32 & 20.26 & 35.51 \\ \hline
			
	\end{tabular}}
	\label{tab:user_study}
\end{table}

\subsection{Ablation study and result analysis}
\label{sec:ablation_study}
\begin{figure*}[htb]
	\centering
	\includegraphics[width=\textwidth]{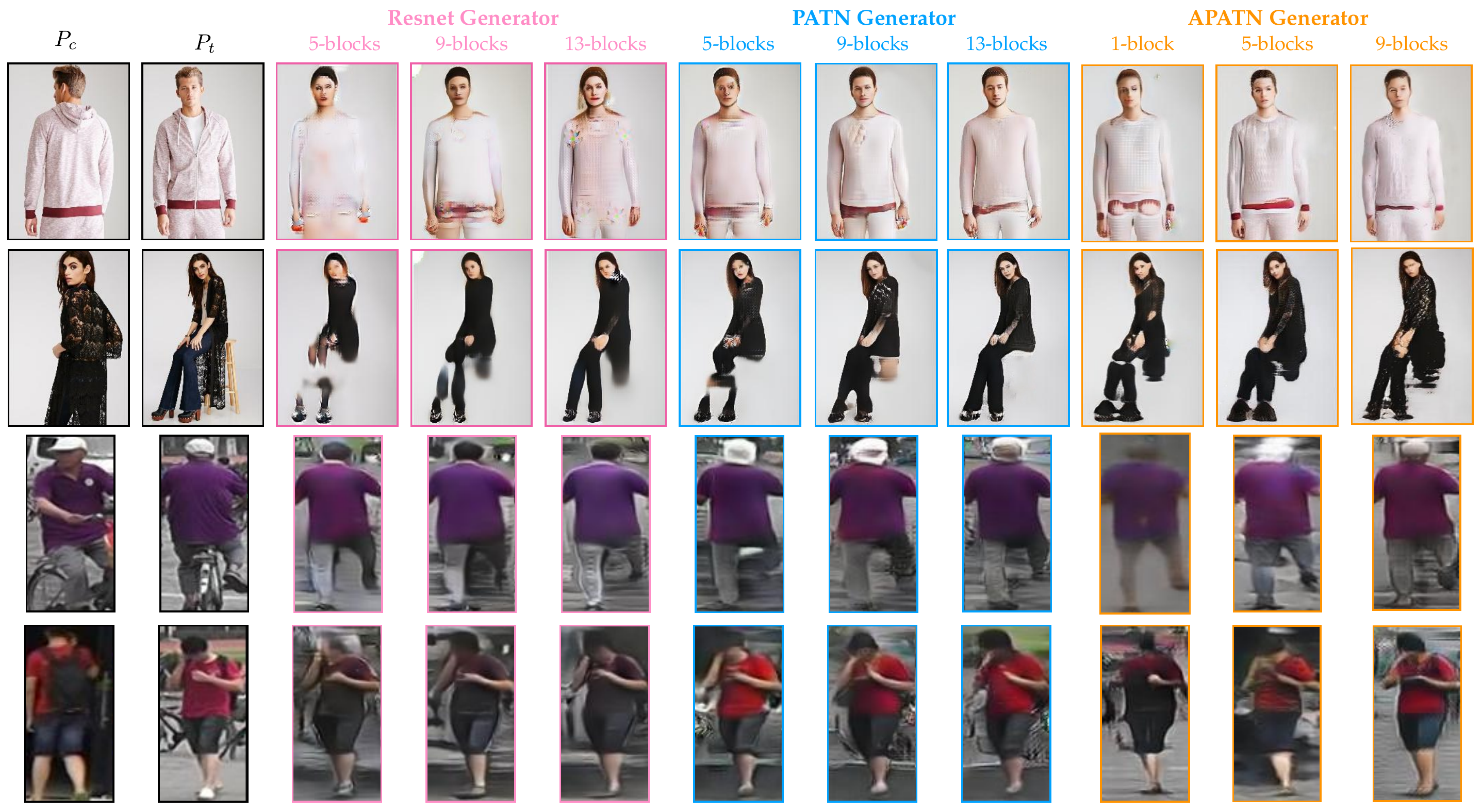}
	\caption{Qualitative results of different types and numbers of the constituent blocks in the generator on Market-1501 and DeepFashion dataset. The images with pink border are generated by Resnet generator \cite{perceptualloss,pix2pix2017-cvpr} containing 5, 9, 13 residual blocks \cite{resnet} from left to right respectively. And the images with blue borders are generated by our \emph{PATN generator} containing 5, 9, 13 PATBs. The images with orange borders are generated by our \emph{APATN generator} containing 1, 5, 9 APATBs.}
	\label{fig:ablation_study}
\end{figure*}

\begin{table*}[htb]
	\centering
	\caption{Quantitative ablation analysis of the progressive generation.}
	\resizebox{\textwidth}{!}{
		\begin{tabular}{|l|cccccc|cccc|}
			\hline
			\multirow{2}{*}{Model} & \multicolumn{6}{c|}{Market-1501}           & \multicolumn{4}{c|}{DeepFashion} \\ \cline{2-11}  & SSIM & IS & mask-SSIM & mask-IS & DS &PCKh &SSIM &IS &DS &PCKh \\ \hline
			Resnet generator (5 blocks) & 0.297 & 3.236 & 0.802 & 3.807 & 0.67 & 0.84 & 0.764 & 2.952 & 0.900 &0.89 \\ 
			Resnet generator (9 blocks)     & 0.301 & 3.077 & 0.802 & 3.862 & 0.69 & 0.87 & 0.767 &3.157 & 0.941 &0.90 \\ 
			Resnet generator  (13 blocks) & 0.300 & 3.134 &0.797 & 3.731 & 0.67 &0.88 &0.766 &3.107 & 0.943 &0.89  \\ \hline
			PATN generator (5 blocks)     & 0.309 & 3.273 & 0.809 &3.870 & 0.69 & 0.91 &0.771 &3.108 & 0.958 &0.93 \\
			PATN generator (9 blocks)  &0.311 & 3.323 & 0.811 & 3.773 & 0.74 & 0.94 &0.773  & 3.209 & 0.976  & 0.96    \\
			PATN generator (13 blocks) & 0.314 & 3.274 &0.808 & 3.797 & 0.75 &0.93 & 0.776 &3.440 & 0.970 & 0.96  \\ \hline
			APATN generator (1 block)     & 0.251 & 3.875 & 0.783	& 3.655 & 0.61 & 0.81 & 0.768 & 3.201	& 0.968 & 0.87 \\
 			APATN generator (5 blocks)     & 0.326	& 3.426 & 0.817	& 3.801 & 0.66 & 0.91  & {0.780} &	3.435 & 0.971 & 0.96 \\
			APATN generator (9 blocks)  & 0.326 & 3.522	& 0.821 & 3.727 & 0.66 & 0.91 & 0.782 & 3.187	& 0.972 & 0.95    \\ \hline
	\end{tabular}}
	\label{ablationstudy_v2}
\end{table*}

The generator of our network -- PATN and APATN have two important design characteristics: one is the carefully designed building blocks -- PATB and APATB, which aim to optimize appearance and pose simultaneously using the attention mechanism; the other is the cascade of building blocks which aims to guide the deformable transfer process progressively. 
In this case, we first carry out two sets of experiments: one is to verify the design of the proposed building blocks, and the other is to exhibit the advantage of the progressive manner by varying the number of building blocks in the PATN or APATN generator. 


\subsubsection{Analysis of the building blocks}

\begin{table*}[htb]
	\centering
	\caption{Quantitative comparison of the ablation study on dissection of PATB and APATB.}
	\resizebox{\textwidth}{!}{
		\begin{tabular}{|l|cccccc|cccc|}
			\hline
			\multirow{2}{*}{Model} & \multicolumn{6}{c|}{Market-1501}           & \multicolumn{4}{c|}{DeepFashion} \\ \cline{2-11}  & SSIM & IS & mask-SSIM & mask-IS & DS &PCKh &SSIM &IS &DS &PCKh \\ \hline
			PATB  & 0.311  & 3.323  & 0.811  & {3.773}  & 0.74   & 0.94  & 0.773  & 3.209 & {0.976}  & {0.96} \\
			PATB w/o add  & 0.305 & 3.650 & 0.805 & 3.899 & 0.73 & 0.92 & 0.769 & 3.265 & 0.970 & 0.93  \\
			PATB w/o cat  & 0.306 & 3.373 & 0.807 & 3.915 & 0.71 & 0.92 & 0.767 & 3.200 & 0.968 & 0.95  \\ \hline
 			APATB & 0.326	& 3.426 & 0.817	& 3.801 & 0.66 & 0.91  & {0.780} &	3.435 & 0.971 & 0.96 \\
			APATB cat$\rightarrow$add & 0.333 & 3.861 & 0.821 & 3.744	& 0.60	& 0.91 & 0.787	& 2.972	&	0.967	& 0.95\\ 
			APATB w/o cat & 0.233 &	4.071 & 0.768 &	3.766 &	0.61 & 0.75 & 0.767 & 3.28 & 0.942 & 0.90  \\ 
			APATB w/o res & 0.255 &	4.073 &	0.783 &	3.660 & 0.57 & 0.78 & 0.773 & 3.35 & 0.955 & 0.91	  \\ \hline
	\end{tabular}}
	\label{tab:quantitative_comparison_pmtn}
\end{table*}

\vspace{1ex}\noindent\textbf{Advantages of PATB and APATB.}~We further explore the advantages of PATB and APATB by replacing them with vanilla residual block \cite{perceptualloss} that results in a generator named \emph{Resnet generator}. A side-to-side qualitative comparison is shown in Fig.~\ref{fig:ablation_study}. 

It seems that PATN generators can always generate images demonstrating much more consistent shape and appearance with its target image. Moreover, the Resnet generator is prone to ignore some indistinguishable but representative appearance information which usually occupies a small portion of the image, and fails to generate correct foreground shape when the target pose is relatively rare. For instance, Resnet generator is likely to ignore the bottom red ring of the sweater in the first row, the white cap in the third row and mislead the T-shirt color as black in the fourth row since a black backpack occludes a large portion of the T-shirt. Besides, the shapes of the sitting girls produced by Resnet generator in the second row are somewhat incomplete as this pose is quite rare in DeepFashion~\cite{DeepFashion}. We assume this should be attributed to the pose attention mechanism which enhances model's abilities in capturing useful features and leveraging them.

Under almost all the quantitative measures shown in Tab.\ref{ablationstudy_v2}, our PATN generator with only 5 PATBs outperforms Resnet generator with all its number of residual blocks configurations. These results clearly demonstrate the advantages of our PATN generator. Similar conclusion can be observed for our APATN.

\begin{figure}[htb]
	\centering
	\includegraphics[width=0.5\textwidth]{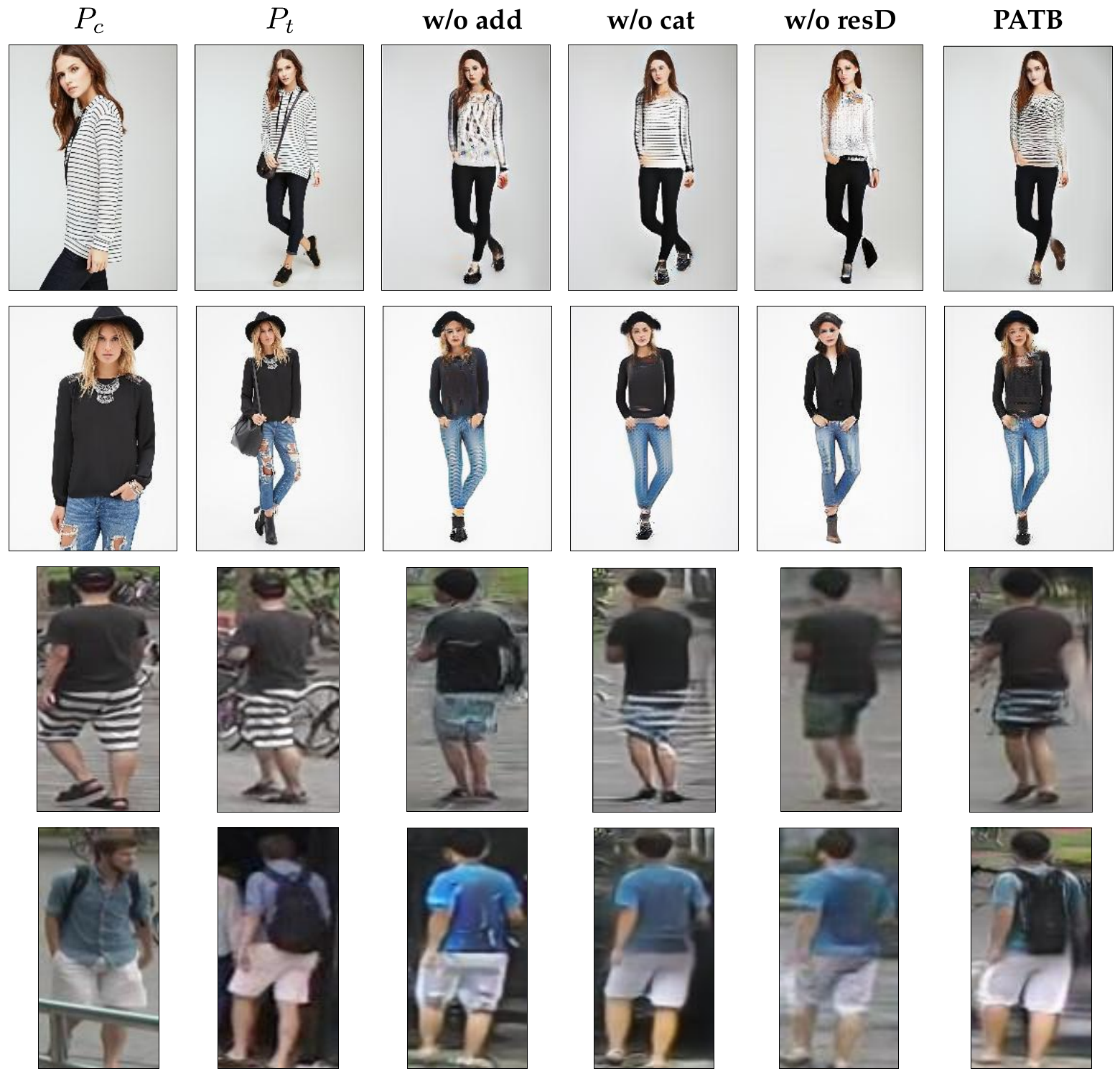}
	\caption{Qualitative results of the ablation study for PATB on Market-1501 and DeepFashion dataset.}
	\label{fig:ablation_study_PATB}
\end{figure}

\vspace{1ex}\noindent\textbf{Dissection of the PATB.}~To investigate the effect of each part of PATB, we conduct experiments by removing the addition (\textbf{PATB w/o add}) and concatenation (\textbf{PATB w/o cat}) operation in every PATB inside the PATN generator (9 blocks). The qualitative and quantitative results are given in Fig.\ref{fig:ablation_study_PATB} and Tab.\ref{tab:quantitative_comparison_pmtn}. It can be observed that by removing any parts of PATB would lead to a performance drop, visually exhibiting a certain degree of color distortion and implausible details. 

\vspace{1ex}\noindent\textbf{Dissection of the APATB.}~We further explore potential alternatives of the APATB, including: 1) changing the concatenation operation in the image code update process to addition operation (\textbf{APATB cat$\rightarrow$add});  2) removing the concatenation operation in the image code update step (\textbf{APATB w/o cat}); 3) discarding the residual block $\text{conv}_{\tau}^S$ used in the pose code update step from APATB (\textbf{APATB w/o res}).


{In Tab.~\ref{tab:quantitative_comparison_pmtn}, we present the quantitative comparisons between the generator equipped with 5 APATBs and those with the alternatives (also 5 blocks). The large performance boost of APATB over \emph{APATB w/o cat} and \emph{APATB w/o res} on most metrics, such as SSIM, mask-SSIM, DS and PCKh, clearly shows that leveraging residual block to conduct the pose code update step and using the concatenation operation during the image code update process, are two very critical design criterion for APATB. The residual block enlarges the receptive field, and plays an important role in updating the pose code for better expression of pose joints, which is beneficial for calculating the pose alignment. Moreover, at the early stage, it is highly challenging for APATB to accurately calculate the alignment, which could result in huge information loss of useful appearance features and leads to unstable training. In this case, the concatenation and addition operation can effectively reduce the information loss and ease the training process. It can be observed that adopting concatenation leads to largely higher DS while slightly lower SSIM than the results of using addition~(\emph{APATB cat$\rightarrow$add}) on both datasets. Hence, we lean on selecting concatenation in our final design of APATB since it achieves consistent improvements on most metrics.}


\begin{figure}[htb]
	\centering
	\includegraphics[width=0.5\textwidth]{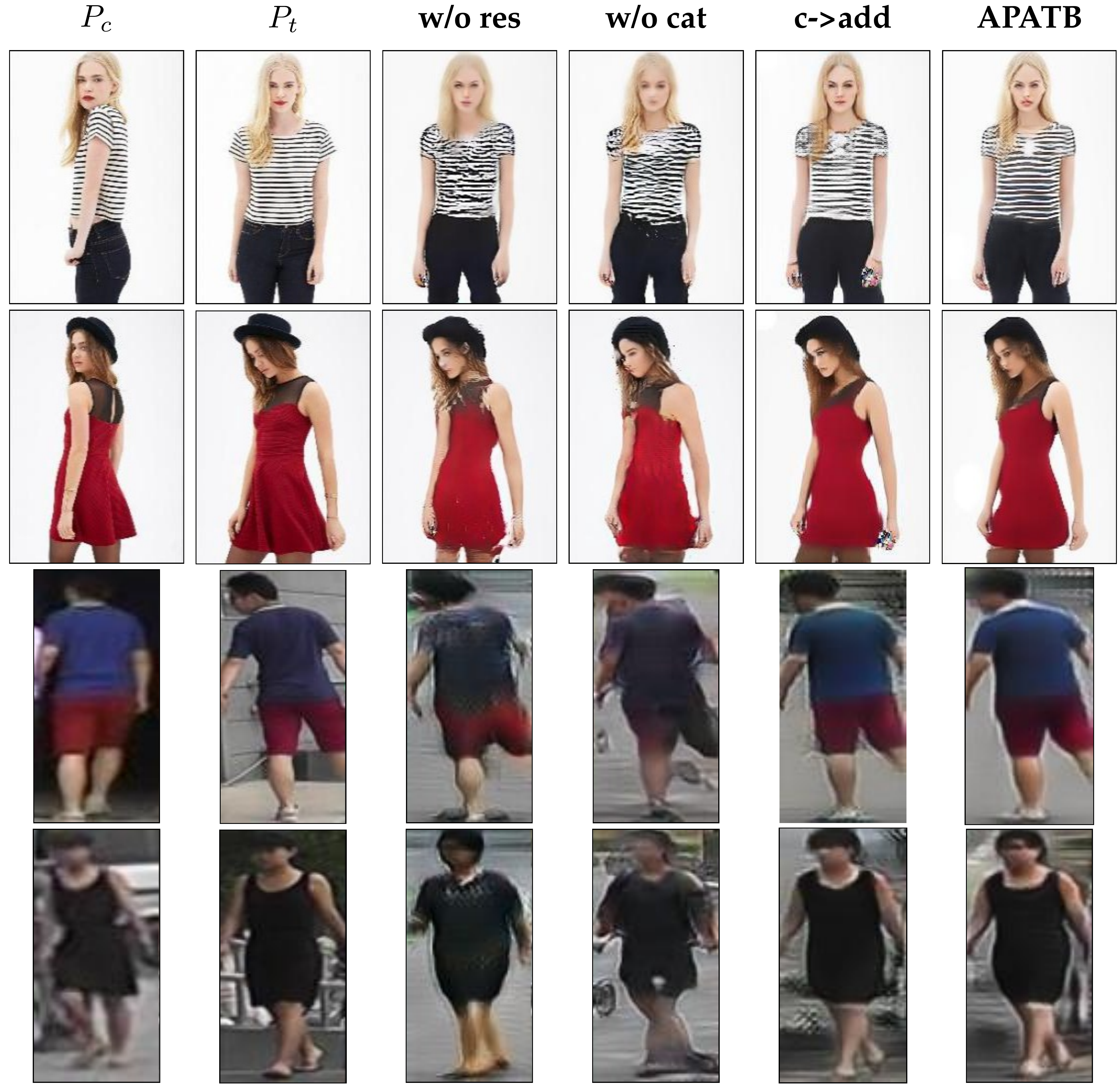}
	\caption{Qualitative results of the ablation study for APATB on Market-1501 and DeepFashion dataset.}
	\label{fig:ablation_study_APATB}
\end{figure}

We also give several representative qualitative results in Fig.~\ref{fig:ablation_study_APATB}. It can be easily read from these results that APATB is superior than the other three competitors in terms of the fidelity of the images and the integrity of the human body. Besides, APATB recovers more details than the other three such as the style of the upper clothes in the first row, the hat in the second row, the collar in the third row and the style of the dress in the last row. Without using residual blocks to update pose codes, the results cannot model the inner pose alignment, leading to incorrect and abnormal human body reconstruction. For example in the fourth row, the legs should cross but the \emph{w/o cat} model produces a person with two feet standing side by side. Without the concatenation operation between the previous image code and the current pose code, there exists an obvious information loss, \emph{e.g.}, the color of the pants is incorrectly reconstructed in the third row. These results visually demonstrate the necessity of the concatenation operation in the image code update step and the residual blocks used in the pose code update step. {As for the choice of concatenation or addition}, the visual results on the DeepFashion dataset show that concatenation tends to preserve more accurate details and exhibits fewer artifacts.

\begin{figure}[htb]
	\centering
	\small
	\includegraphics[width=0.5\textwidth]{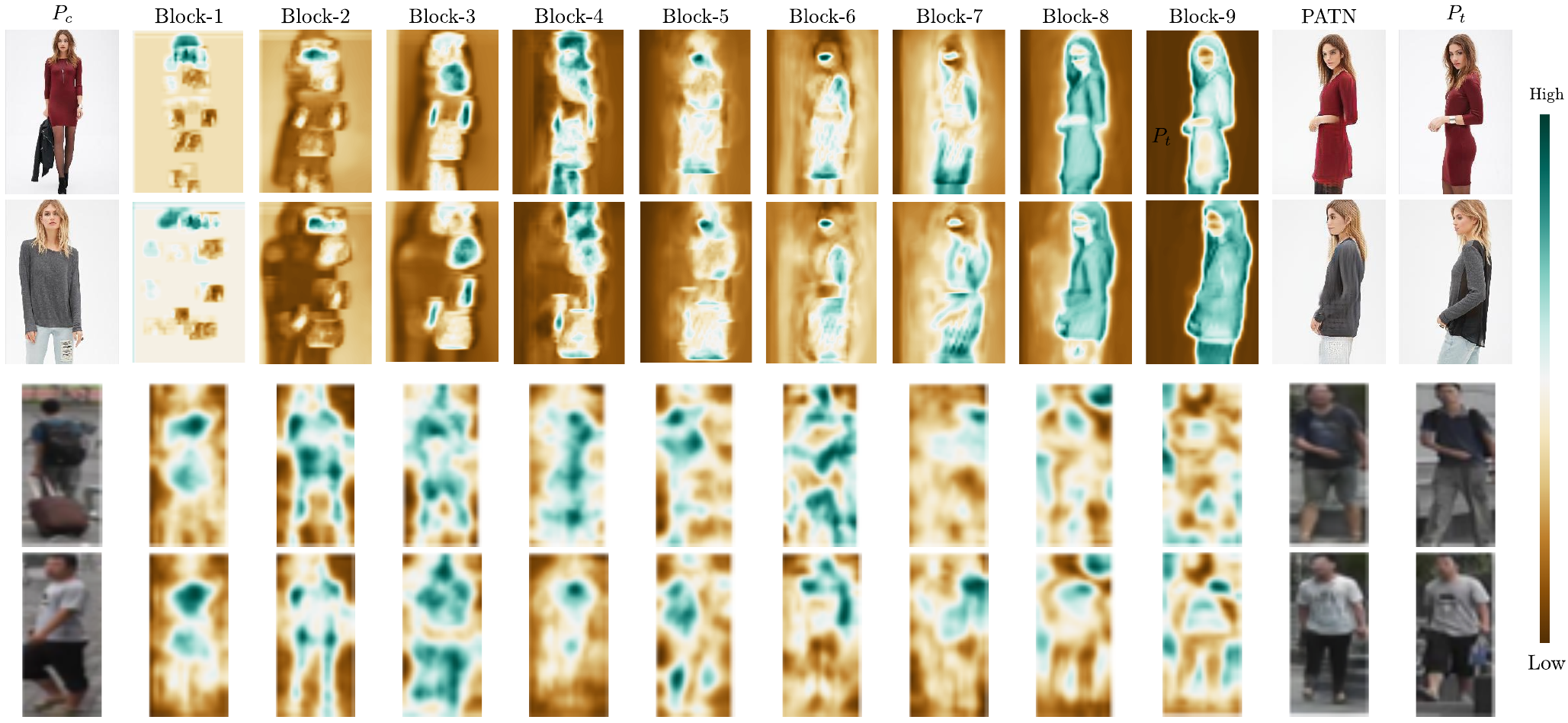}
	\caption{Visualization of the attention masks in PATBs. ``Block-1'' denotes the attention mask of the first PATB and ``Block-2$\sim$9'' likewise.} 
	\label{fig:attblock_vis_patn}
\end{figure}

\vspace{1ex}\noindent\textbf{Visualization of the attention masks in PATBs and APATBs.}~To get an intuitive understanding on how the PATBs work during the pose transfer process, we visualize the attention masks of the cascaded PATBs in Fig.\ref{fig:attblock_vis_patn}. For PATB, the initial several masks (first three columns of masks) are some sort of blending of condition pose and target pose. As the condition pose is transferred towards the target pose, the regions needed to be adjusted are shrunk and scattered, reflected in the middle four mask columns. Eventually, the last two mask columns show that the attention is {focused on the foreground person for the appearance refinement on the DeepFashion dataset while transformed to background refinement on the Market-1501 dataset. Such different behaviors of the last two PATBs demonstrate the flexibility of the PATN generator to adjust its power to overcome different challenges posed by different datasets:} {rich appearance in the DeepFashion dataset and complex backgounds in the Market-1501 dataset}. 

\begin{figure}[htbp]
	\centering
	\small
	\includegraphics[width=0.5\textwidth]{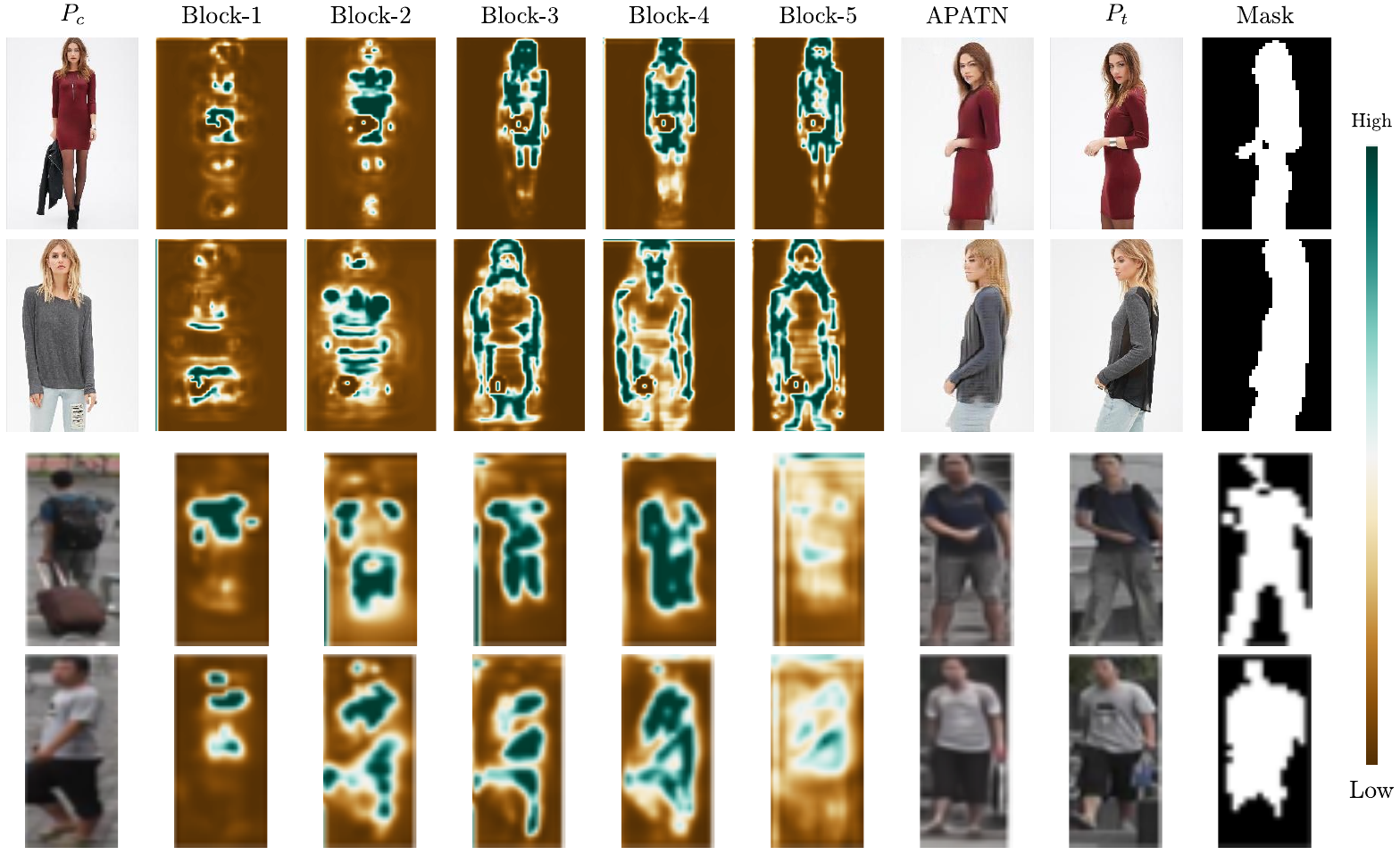}
	\caption{Visualization of the attention masks in APATBs. ``Block-1'' denotes the attention mask of the first APATB and ``Block-2$\sim$5'' likewise.} 
	\label{fig:attblock_vis_APATN}
\end{figure}

{We further present the visualization for the attention masks of the APATBs in Fig.~\ref{fig:attblock_vis_APATN}. To figure out how APATB aligns target pose with the condition pose, we extract a shape mask of the person via a human parser~\cite{HuangXBWB19} and visualize the attention mask of the pixels inside the shape mask. As shown in Fig.~\ref{fig:attblock_vis_APATN}, the attention masks of the cascading APATBs gradually cover the related foreground regions, which are leveraged to generate the target image. We observe a different behavior of the last APATB for different datasets, which is similar to the phenomenon observed in the last two PATBs. The last APATB largely focuses on the person for appearance refinement for the DeepFashion daset. However, it produces a relatively scattered attention mask in both the foreground and the background for the Market-1501 dataset. This is helpful to naturally embed the person into the complex background. It is worth noting that the APATB can adaptively ignore the unrelated regions. For example, in the first row of Fig.~\ref{fig:attblock_vis_APATN}, the APATBs pay less attention to the legs since the target images does not involve the legs. Similarly, in the second row of Fig.~\ref{fig:attblock_vis_APATN}, the APATBs pay less attention to the center of the torso, since the target pose is under side view.} 




\subsubsection{Analysis of the progressive generation}

To analyze the influence of the progressive generation process, we further conduct a set of experiments based on PATN with its number of PATBs set to 5, 9, 13, respectively. Besides, we also show the progressive generation process also benefit the Resnet generator by setting the number of the residual blocks to 5, 9, 13, respectively. Quantitative results are shown in Tab.~\ref{ablationstudy_v2} and the qualitative results are show in Fig.~\ref{fig:ablation_study}. It is evident that while the block number is increasing from 5 to 9, the performances of both PATN generator and Resnet generator steadily improve. 
Further increasing the number of PATBs to 13 can marginally boost the performance. This phenomenon reflects that increasing the number of PATBs will ease the transfer process and give results with less artifacts. For a tradeoff, we chose 9 PATBs as default for better efficiency. 

Besides, we also investigate the effect of the progressive generation process of the APATN generator. Since APATB generator is computationally more expensive than PATN generator, we conduct experiments based on APATN with its block number set to 1, 5, 9, respectively. Both qualitative and quantitative results of the APATN generator with only one APATB are noticeably worse than those with 5 or 9 APATBs. However, the difference between the APATN generator with 5 APATBs and that with 9 APATBs is minor. Considering the tradeoff between the computational cost and performance, we opt to using 5 APATBs as the default configuration for our APATN generator.

\subsection{Failure cases}

\begin{figure}
	\centering
	\includegraphics[width=0.48\textwidth]{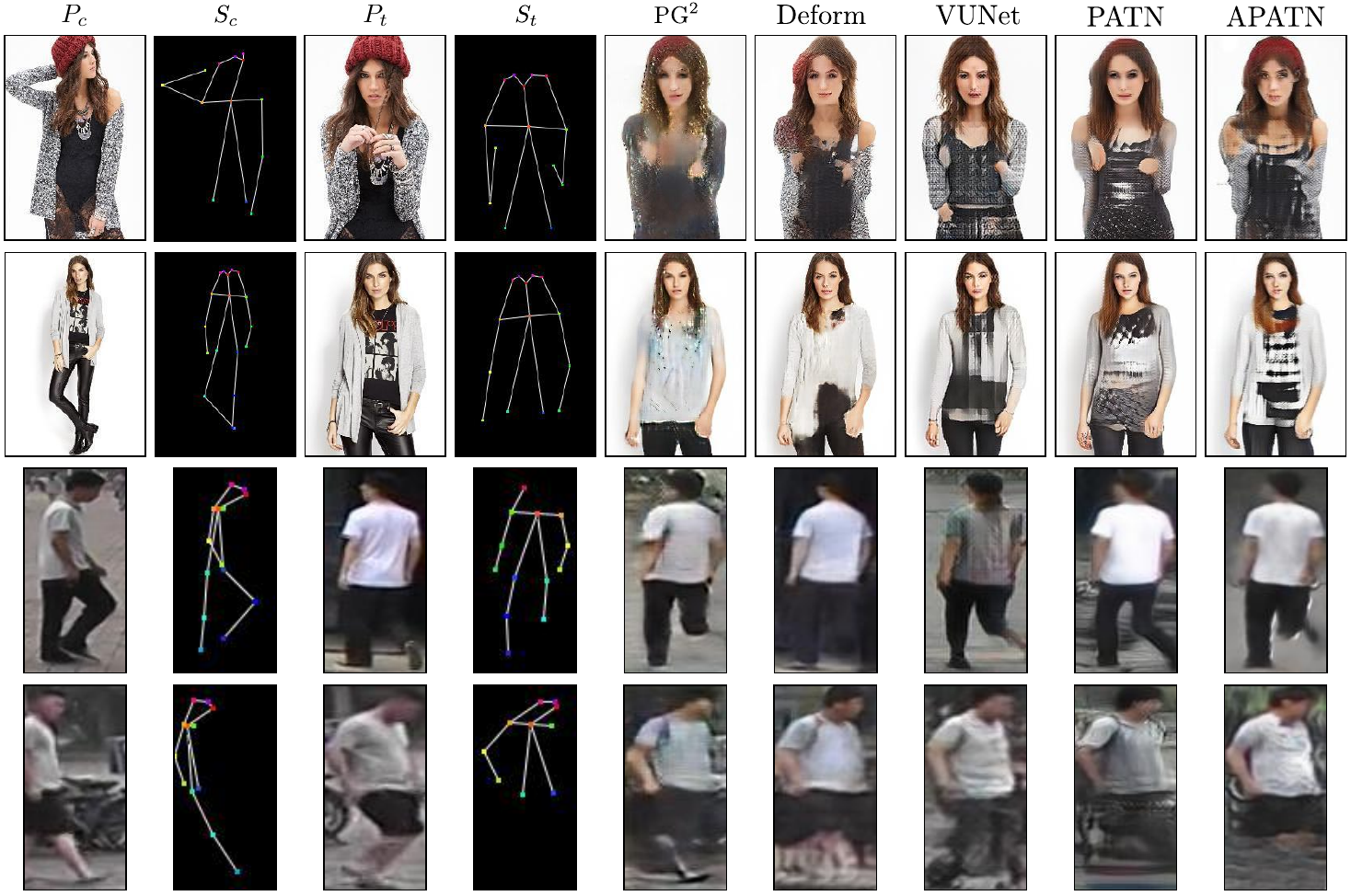}
	\caption{Illustration of several failure cases on the DeepFashion and Market-1501 dataset.}
	\label{fig:failure}
\end{figure}

Though APATN and PATN achieve excellent performance in generating realistic person images, they may fail in some challenging cases, such as rare poses, complex appearance, incomplete poses. In Fig.~\ref{fig:failure}, we demonstrate several typical examples, which we believe could provide nutrients and directions for further researches. For example, for the woman in the first row, it is hard for the all the models to generate plausible hands since such posture is quite rare in the dataset. In the second row, the upper clothes of the women are so complex that different models behave differently. $\text{PG}^2$ and $\text{Deform}$ tend to reconstruct the apparent white sweater and ignore the inner black shirt with complex decorative pattern. $\text{VUNet}$ and our PATN and APATN struggle to reconstruct the shirt but eventually fail to reach such goal. For the third row, our PATN generates plausible background scenes while APATN generates blurry backgrounds. This sample reveals that APATN has comparatively weaker ability to reconstruct backgrounds, given no explicit guidance for the background. \edit{Future works may follow~\cite{unseenpose} to treat foreground and background separately and then blend together to alleviate the background reconstruction problem.} Besides, for the last rows, the incomplete pose leads to incomplete human body of all models. These cases expose that current keypoint-based pose transfer models still have some challenges to overcome, including rare and incomplete poses. These challenges require extensive future study.



\section{Application to person re-identification}

\begin{table*}[htbp]
	\centering
	\caption{The ReID results on Inception-v2 and ResNet-50 using images generated by different methods. \textit{None} means no generative model is employed. * denotes the results when we randomly select target poses from $M^R_p$ for data augmentation. $\dagger$ indicates the ResNet-50 trained with Circle loss~\cite{circleLoss}.}
	\resizebox{\textwidth}{!}{
		\begin{tabular}{|l|l|cccccccccc|cc|}
			\hline
			\multirow{2}{*}{Augmentation Model} &
			\multirow{2}{*}{Backbone Model} &
			\multicolumn{10}{c|}{Portion $p$ of the real images} &
			\multicolumn{2}{c|}{Aug. ratio}\\ 
			\cline{3-14}  & & 0.1 &0.2 &0.3 &0.4 &0.5 &0.6 &0.7 &0.8 &0.9 &1.0 &2 &3 \\\hline
			None &Inception-v2 &5.7 &16.6 &26.0 &33.5 &39.2 &42.2 &46.5 &49.0 &49.5 &52.7 &- &- \\
			PATN &Inception-v2 &42.3 &43.6 &45.7 &46.7 &48.0 &48.5 &49.1 &50.7 &51.6 &52.7 &56.6 &57.1 \\\hline
			None &ResNet-50 &9.2 &27.6 &41.5 &50.3 &56.2 &58.8 &61.2 &62.7 &63.8 &65.3 &- &- \\ 
			VUNet\cite{vunet} &ResNet-50  &49.8 &51.7 &53.7 &54.5 &56.6 &58.4 &59.4 &61.3 &62.9 &65.3 &63.9 &64.1\\ 
			Deform\cite{DeformableGAN} &ResNet-50 &51.9 &53.9 &55.4 &56.1 &57.6 &59.4 &60.5 &62.2 &63.5 &65.3 &64.2 &64.6 \\ \hline
			PATN &ResNet-50    &52.6 &54.5 & 56.5 &56.6 &60.3 & 60.9 & 62.1 & 63.3 & 64.8 &65.3 & 65.3 & 65.7 \\
			PATN* &ResNet-50    & 53.3 & 55.9 &56.0 & 57.3 &58.8 &60.4 &60.7 &63.1 &64.5 &65.3 &65.1 &65.4 \\ 
			APATN* &ResNet-50     &52.7 &55.7 & 56.1 & 57.1 & 58.9 & 60.2 & 61.4 & 63.4 & 64.3 & 65.3 & 65.1 & 65.5 \\ \hline
			None &ResNet-50 $\dagger$ &6.4 &28.8 &50.9 &63.8 &73.1 &76.2 &80.0 &81.3 &82.3 & 84.7 &- &- \\ 
			PATN* &ResNet-50 $\dagger$ &60.5 &66.7 &69.3 &73.2 &74.8 &77.9 &79.4 &81.6 &83.1 &84.7 & 84.3 & 83.5\\ 
			APATN* &ResNet-50 $\dagger$ &60.7 &66.2 &69.9 &72.7 &75.3 &77.4 &80.1 &81.5 &82.8 &84.7 &84.1 & 83.7 \\ \hline
	\end{tabular}}
	
	\label{tab:reid_aug}
\end{table*}

A good person pose transfer method can generate realistic-looking person images to augment the datasets of person-related vision tasks, which might result in better performance especially in the situation of insufficient training data. As person re-identification (re-ID) \cite{personreid} is increasingly promising in real-world applications and has great research significance, we choose to evaluate the performance of our method in augmenting the person re-ID dataset. \edit{Specifically, we use the mainstream person re-ID dataset Market-1501 \cite{Market1501} and three advanced re-ID networks, Inception-v2 \cite{inception}/ResNet-50 \cite{resnet} trained with cross-entropy loss, and ResNet-50$\dagger$  trained with circle loss~\cite{circleLoss} , as our test bed.}


We first randomly select a portion $p$ of the real Market-1501 dataset as the reduced training set, denoted as $\mathcal{M}^{R}_{p}$, where at least one image per identity is preserved in $\mathcal{M}^R_{p}$ for better person identity variety. Each unselected training image $\widetilde{I}_i$ is replaced by a generated image that transfers from a randomly chosen image with the same identity in $\mathcal{M}^R_{p}$ to the pose of $\widetilde{I}_i$.
Consequently, an augmented training set $\mathcal{M}^A_{p}$ is formed from all these generated images and $\mathcal{M}_{p}^R$. 
The selected portion $p$ varies from 10\% to 90\% at intervals of 10\%. We follow the training and testing protocols of the Open-Reid Framework\footnote{\url{https://github.com/Cysu/open-reid}} to train the ResNet-50 and Inception-v2 on each of the nine reduced training set $\mathcal{M}^R_{p}$  and the nine augmented training set $\mathcal{M}^A_{p}$ respectively.

We argue that our experiment scheme is well-suited for assessing the usefulness of our image generation method in boosting the performance of re-ID via data augmentation due to the following two reasons: 1) the original training set reduction leads to a degree of data insufficiency, offering an opportunity to boost the performance via data augmentation and giving the performance \emph{lower bound}; 2) the reference results utilize all the real images including all the ground truth of the generated images, thus could give the theoretical performance \emph{upper bound}. Besides, the gaps between the upper and lower bounds could measure the maximum potential of data augmentation in boosting re-ID performance. 

Tab.\ref{tab:reid_aug} shows that whenever there's a performance gap, there is undoubtedly a performance boost owing to our data augmentation. And the performance gain is much more significant if the performance gap is relatively large. A natural question is, what the performance would be when further augmenting the whole real training set. To investigate, we augment the training dataset of Market-1501 by generating one/two samples per image, whose target poses are randomly selected from the whole dataset, thus doubles/triples the size of the original training set. The results on these two augmented datasets are added to Tab.\ref{tab:reid_aug} (the right part). In a nutshell, the trends of performance gain by adding more generated images are roughly in accordance with those by adding more real images. It can be seen that the real data augmentation for Inception-v2 model could get nearly linear improvements even in cases of near sizes of the whole real data set. Therefore, doubling or tripling the training set continuously improves the performance considerably. On the other hand, real data augmentation for ResNet-50 model tends to saturate in cases of near sizes of the whole real data set, hence doubling or tripling the size fails to improve the performance further. \edit{It is worth noting that enlarging the full Market-1501 dataset with the generated images even leads to slightly worse results for the strong baseline ResNet50$\dagger$, although the generated images indeed contribute to significant improvements over the reduced dataset. It reveals that the generated images play a key role when the available data is scarce. However, the gap between the generated images and real images may bias the feature learning progress, and thus leads to 
less promising performance. }

We further compare our method to several existing person image generators~\cite{vunet, DeformableGAN} under the same setting for re-ID data augmentation. As shown in Tab.\ref{tab:reid_aug}, our method achieve consistent improvements over previous works for different portion $p$ of the real images, suggesting the proposed method can generate more realistic human images and be more effective for the re-ID task. We also present the re-ID performance by randomly selecting the target poses from the whole dataset, which can better suit the practical applications. As shown in Tab.\ref{tab:reid_aug}, there is not an obvious performance difference between the two settings, further demonstrating that the proposed framework is robust to various poses. 

In terms of this application, APATN performs slightly worse than PATN, which might attribute to tendency of APATN to over-simplify the background scenes of the generated images of the Market-1501 dataset. Although the Market-1501 dataset contains many images for a single person and various scenes, it is reasonable that different images of a single person are very possibly captured under the same scene, which indicates that background may be also an important factor for achieving superior performance on the ReID task. This discussion also indicates one of our future work to take background scenes into account while conducting pose transfer.



\section{Conclusion}
In this paper, we propose a progressive pose attention transfer network, with several cascading transfer blocks, to deal with the challenging pose transfer task. Each transfer block optimizes appearance and pose simultaneously using the attention mechanism, thus guides the deformable transfer process progressively. We introduce two types of transfer blocks, namely Pose-Attentional Transfer Block~(PATB) and Aligned Pose-Attentional Transfer Block~(APATB). Compared to PATB, APATB explicitly aligns two poses through evaluating a similarity matrix and provides richer guidance for the sample-and-put operation, thus leading to better appearance quality. Visualization for the attention masks clearly demonstrates the progressive transfer process, making our network more interpretable. Compared to previous works our network exhibits superior performance in both subjective visual realness and objective quantitative scores while possesses higher computational efficiency and less model complexity. 

\section*{Acknowledgment} 
This work is supported by National Natural Science Foundation of China (61733007). This work also gets support from the National Program for Support of Top-notch Young Professionals and the Program for HUST Academic Frontier Youth Team 2017QYTD08, given to Dr. Xiang Bai.

{\small
		\bibliographystyle{IEEEtran}
		\bibliography{egbib}
}
\ifCLASSOPTIONcaptionsoff
  \newpage
\fi



%

%

\begin{IEEEbiography}[{\includegraphics[width=\textwidth]{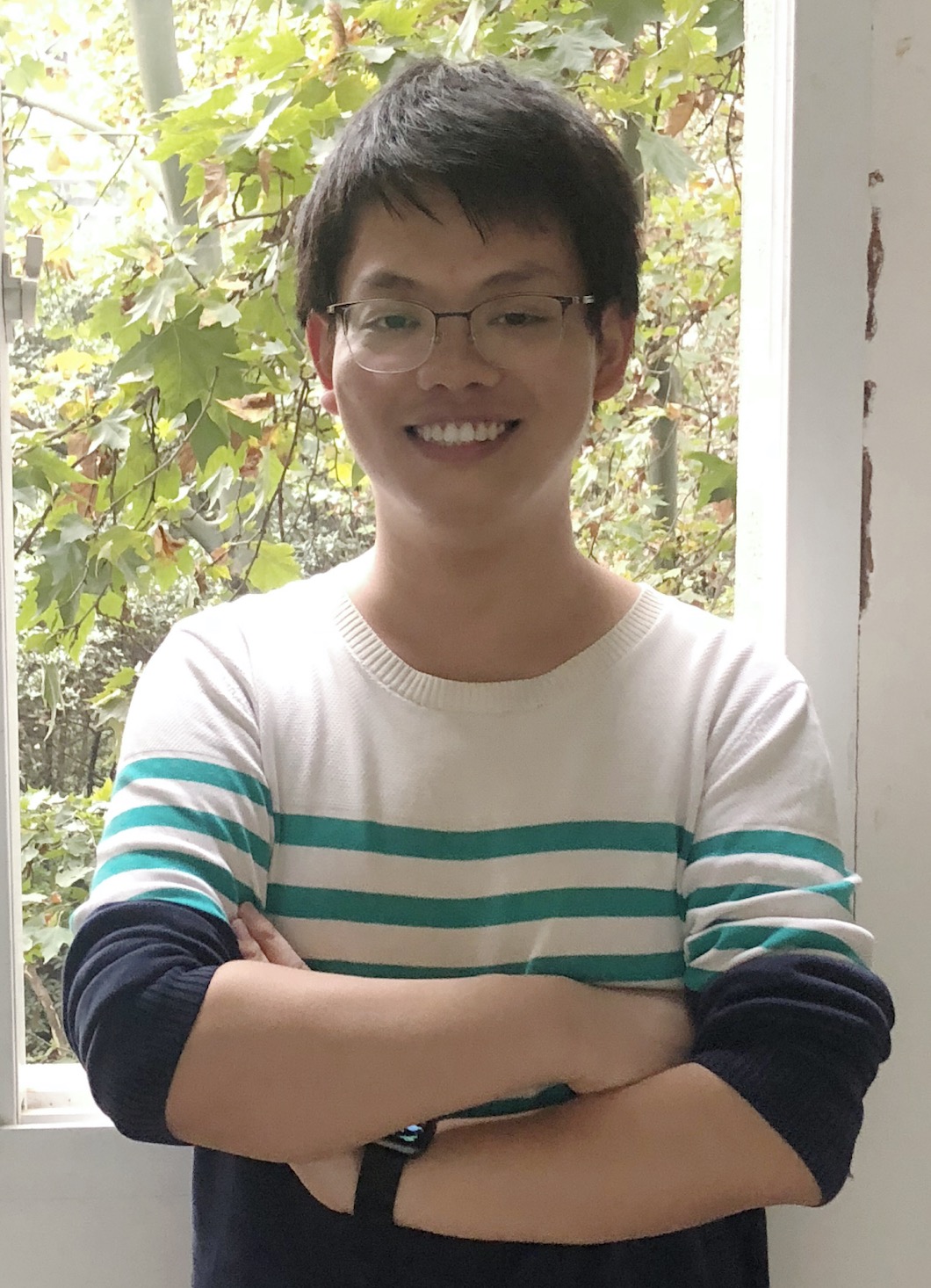}}]{Zhen Zhu} received his B.S. degree from the School of Computer Science and Technology, Huazhong University of Science and Technology (HUST), China in 2017. He received his M.S. degree from the School of Electronic Information and Communications, HUST in 2020. He is currently a PhD student in the Department of Computer Science, University of Illinois at Urbana-Champaign. His main research interests include image generation, semantic segmentation and object detection.
\end{IEEEbiography}

\begin{IEEEbiography}[{\includegraphics[width=\textwidth]{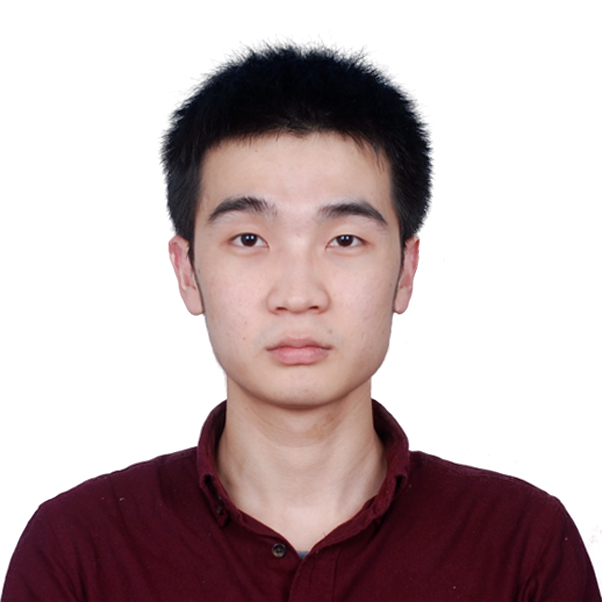}}]{Tengteng Huang} received his B.S. degree from the School of Computer Science and Technology, Huazhong University of Science and Technology (HUST), China in 2017. He received his M.S. degree from the School of Electronic Information and Communications, HUST in 2020. His main research interests include image generation, image retrieval and 3D object detection.
\end{IEEEbiography}

\begin{IEEEbiography}[{\includegraphics[width=\textwidth]{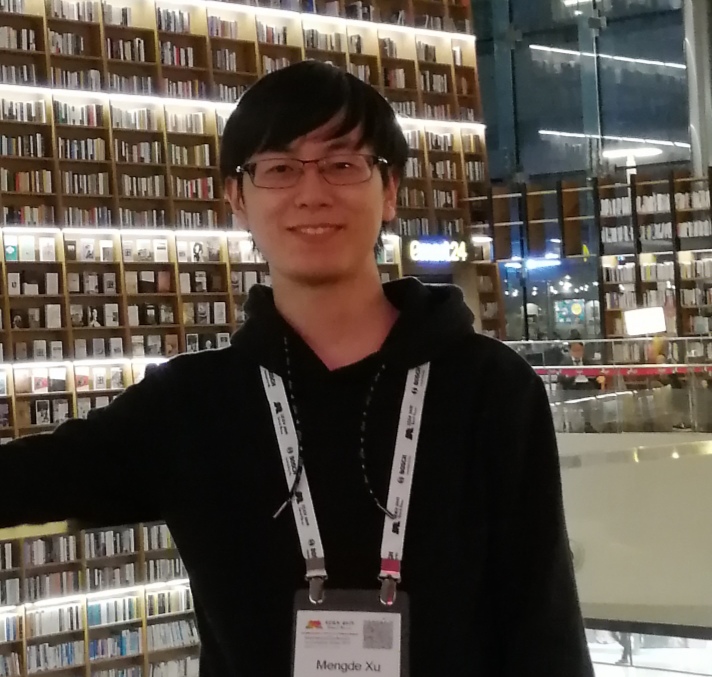}}]{Mengde Xu} received his B.S. degree from the School of material science and engineering, Huazhong University of Science and Technology (HUST), China in 2018. He is currently a PhD student with the School of Electronic Information and Communications, HUST. His main research interests include semantic segmentation, image generation.
\end{IEEEbiography}

\begin{IEEEbiography}[{\includegraphics[width=\textwidth]{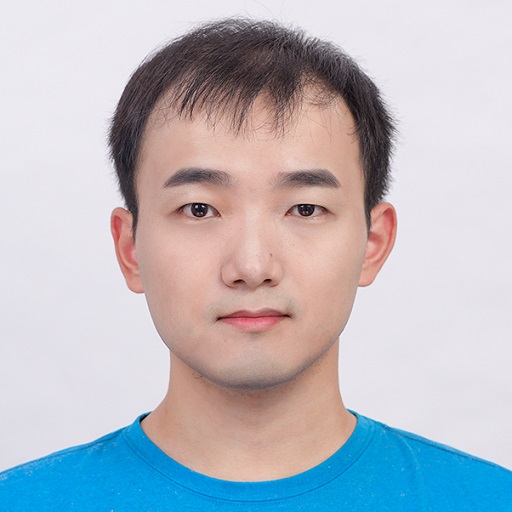}}]{Baoguang Shi} received his B.S., M.S., Ph.D. degree from the School of Electronic Information and Communications, Huazhong University of Science and Technology, Wuhan, China. He was an intern at Microsoft Research Asia in 2014, and a visiting student at Cornell University from 2016 to 2017. He is now a Senior Researcher at Microsoft, Redmond. His research interests include scene text detection and recognition and 3D shape recognition.
\end{IEEEbiography}

\begin{IEEEbiography}[{\includegraphics[width=\textwidth]{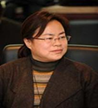}}]{Wenqing Cheng} received her B.S., M.S., Ph.D. degree from the School of Electronic Information and Communications, Huazhong University of Science and Technology, Wuhan, China, in 1985, 1988, and 2005, respectively. She is currently a Professor with the School of Electronic Information and Communications, HUST.
\end{IEEEbiography}

\begin{IEEEbiography}[{\includegraphics[width=\textwidth]{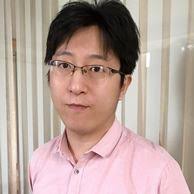}}]{Xiang Bai} received his B.S., M.S., and Ph.D. degrees from the Huazhong University of Science and Technology (HUST), Wuhan, China, in 2003, 2005, and 2009, respectively, all in electronics and information engineering. He is currently a Professor with the School of Artificial Intelligence and Automation, HUST. He is also the Vice-director of the National Center of Anti-Counterfeiting Technology, HUST. His research interests include object recognition, shape analysis, scene text recognition and intelligent systems. He serves as an associate editor for Pattern Recognition , Pattern Recognition Letters, Neurocomputing and Frontiers of Computer Science.
\end{IEEEbiography}




\end{document}